\documentclass[sigconf]{acmart}
\AtBeginDocument{%
  \providecommand\BibTeX{{%
    \normalfont B\kern-0.5em{\scshape i\kern-0.25em b}\kern-0.8em\TeX}}}

\copyrightyear{2023}
\acmYear{2023}
\setcopyright{acmlicensed}\acmConference[KDD '23]{Proceedings of the 29th ACM SIGKDD Conference on Knowledge Discovery and Data Mining}{August 6--10, 2023}{Long Beach, CA, USA}
\acmBooktitle{Proceedings of the 29th ACM SIGKDD Conference on Knowledge Discovery and Data Mining (KDD '23), August 6--10, 2023, Long Beach, CA, USA}
\acmPrice{15.00}
\acmDOI{10.1145/3580305.3599917}
\acmISBN{979-8-4007-0103-0/23/08}

\usepackage{caption}
\usepackage{subcaption}
\usepackage{booktabs}
\usepackage{multirow}
\usepackage{enumitem}
\usepackage{array}
\usepackage{subcaption}
\usepackage{amsmath}

\usepackage{pifont}
\usepackage{subcaption}
\usepackage{balance}
\newcommand{\cmark}{\ding{51}}
\newcommand{\xmark}{\ding{55}}

\begin{document}

\title[Towards Suicide Prevention from Bipolar Disorder with Temporal Symptom-Aware Multitask Learning]{Towards Suicide Prevention from Bipolar Disorder\\with Temporal Symptom-Aware Multitask Learning}

\author{Daeun Lee}
\affiliation{%
  \institution{Dept. of Applied Artificial Intelligence}
  \institution{Sungkyunkwan University}
%  \streetaddress{03063}
  \city{Seoul}
  \country{Republic of Korea}
}
\email{delee12@skku.edu}

\author{Sejung Son}
\affiliation{%
  \institution{Dept. of Applied Artificial Intelligence}
  \institution{Dept. of Human-AI Interaction}
  \institution{Sungkyunkwan University}
%  \streetaddress{03063}
  \city{Seoul}
  \country{Republic of Korea} %
}
\email{maze0717@g.skku.edu}

\author{Hyolim Jeon}
\affiliation{%
  \institution{Dept. of Applied Artificial Intelligence}
  \institution{Sungkyunkwan University}   
%  \streetaddress{03063}
  \city{Seoul}
  \country{Republic of Korea}
}
\email{gyfla1512@g.skku.edu}

\author{Seungbae Kim}
\affiliation{%
 \institution{Computer Science and Engineering}
 \institution{College of Engineering}
   \institution{University of South Florida}   
 \city{Tampa}
 \state{Florida}
 \country{USA}
 }
\email{seungbae@usf.edu}

\author{Jinyoung Han}
\authornote{Corresponding author.}
\affiliation{%
  \institution{Dept. of Applied Artificial Intelligence}
  \institution{Dept. of Human-AI Interaction}
    \institution{Sungkyunkwan University}   
%  \streetaddress{03063}
  \city{Seoul}
  \country{Republic of Korea}
}
\email{jinyounghan@skku.edu}

\renewcommand{\shortauthors}{Daeun Lee, Sejung Son, Hyolim Jeon, Seungbae Kim, \& Jinyoung Han}

\renewcommand{\UrlFont}{\ttfamily\small}
\newcommand{\jyhan}[1]{\textcolor{blue}{\textbf{jyhan:#1}}}
\newcommand{\sbkim}[1]{\textcolor{red}{\textbf{sbkim:#1}}}
\newcommand{\sjson}[1]{\textcolor{purple}{\textbf{sjson:#1}}}
\newcommand{\delee}[1]{\textcolor{orange}{\textbf{delee:#1}}}
\newcommand{\hljeon}[1]{\textcolor{pink}{\textbf{hljeon:#1}}}
\newcommand{\hyolim}[1]{\textcolor{green}{\textbf{hyolim:#1}}} %need_check

\begin{abstract}
Bipolar disorder (BD) is closely associated with an increased risk of suicide. However, while the prior work has revealed valuable insight into understanding the behavior of BD patients on social media, little attention has been paid to developing a model that can predict the future suicidality of a BD patient. Therefore, this study proposes a multi-task learning model for predicting the future suicidality of BD patients by jointly learning current symptoms. We build a novel BD dataset clinically validated by psychiatrists, including 14 years of posts on bipolar-related subreddits written by 818 BD patients, along with the annotations of future suicidality and BD symptoms. We also suggest a temporal symptom-aware attention mechanism to determine which symptoms are the most influential for predicting future suicidality over time through a sequence of BD posts. Our experiments demonstrate that the proposed model outperforms the state-of-the-art models in both BD symptom identification and future suicidality prediction tasks. In addition, the proposed temporal symptom-aware attention provides interpretable attention weights, helping clinicians to apprehend BD patients more comprehensively and to provide timely intervention by tracking mental state progression.

\end{abstract}



%%


\maketitle

\section{Introduction}
\begin{figure}[t]
    \centering
    \includegraphics[width=\linewidth]{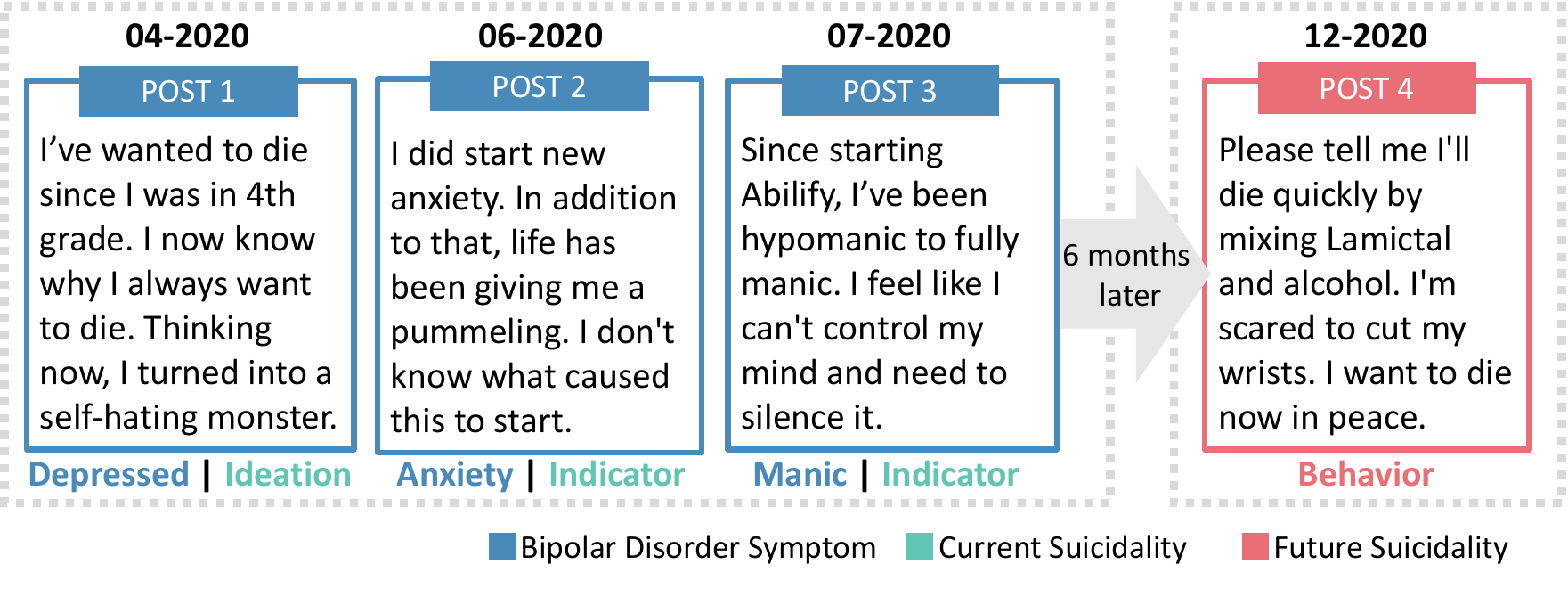}
    \caption{An example of a Reddit user who wrote posts about his/her mental illness on a bipolar disorder-related subreddit and then revealed suicidality 6 months later.}
    \label{fig:user}
    \vspace{-0.2in}
\end{figure}

Suicide is a severe health concern worldwide. According to the OECD, 14.1 per 100,000 people die yearly from suicide in the United States\footnote{\url{https://data.oecd.org/healthstat/suicide-rates.htm}}. Unfortunately, most suicides have been committed by individuals with mental illness~\cite{stack2014mental}. Particularly, people living with bipolar disorder~(BD) are more vulnerable to suicide than people with other psychiatric disorders~\cite{rihmer2002bipolar,ilgen2010psychiatric}. It has been reported that the suicide rate for BD patients is up to 30 times higher than that of the general population~\cite{pompili2013epidemiology}, and suicide fatalities occur in 10-20\% of adults who suffer from BD~\cite{geddes2013treatment}. 

With increasing importance in understanding and analyzing BD patients~\cite{rihmer2002bipolar}, recently, there has been an effort to analyze distinct behavioral characteristics of BD patients and assess their mental states~\cite{coppersmith2014quantifying, cohan2018smhd, vsnajder2018not} using social media data where they share their daily lives and emotions~\cite{kim2020deep, kang2022experiencing}. However, while the prior work has revealed valuable insights into understanding the behavior of BD patients revealed on social media, little attention had been paid to developing a model that can predict the future suicidality of a BD patient. Although a few studies have proposed methods to identify the current risk of suicide in a given social media post~\cite{lee2022detecting, izmaylov2023combining, lee2020cross}, suicidal ideation can often quickly lead to an actual attempt, thereby making them ineffective in preventing suicide~\cite{bryan2016importance,bryan2018predictors,klonsky2016suicide,nock2008cross}; hence, exploring the BD's risk factors that can lead to suicide ideation for predicting future suicidality is crucial. 
Therefore, this paper aims to predict the future suicidality of BD patients based on their mood symptoms history revealed in their past social media data, which has not been thoroughly investigated.

To this end, we first create a novel BD dataset clinically validated by psychiatrists, including future suicidality and bipolar symptoms. Here, we focus on which bipolar symptoms users have, rather than what diagnosed bipolar types they have, because a transdiagnostic approach helps improve understanding of comorbidity, enabling proper interventions than a diagnostic approach~\cite{gruber2008transdiagnostic}. BD is a mood disorder characterized by manic and depressive episodes where two phases show a recurrent pattern that appears and increases over a while, but the following important attribute is not easily considered; many psychological processes are shared in various diagnoses ~\cite{gruber2008transdiagnostic,dalgleish2020transdiagnostic}, e.g., anxiety can appear in both depression and manic episodes of BD~\cite{bowden2007development}. Figure~\ref{fig:user} illustrates example posts written by an individual with BD gradually leading to suicide. Therefore, timely tracking of mood symptoms that affect future suicidality is inevitable for early intervention, leading to shorter treatment periods and better prognosis in BD patients. However, with the rapid mood swings in BD and limited self-reports from patients, there is a significant gap in understanding the actual path of mood changes between the real world and the conventional clinical setting where clinicians can only see patients under limited conditions and rely on the subjective words of the patients~\cite{harvey2022natural}. Hence, using real-world data derived from patient reports at the nonclinical scene, such as social media, is helpful to understand better BD symptoms~\cite{jagfeld2021understanding, yoo2019semantic, mandla2017being, sahota2020bipolar}. 

Particularly, we collect social media posts from BD communities on Reddit. We then labeled our dataset, which contains 7,592 posts published by 818 users, following the guidelines outlined in the Columbia Suicide Severity Rating Scale (C-SSRS)~\cite{posner2011columbia} and Bipolar Inventory of Symptoms Scale (BISS)~\cite{bowden2007development} for annotating suicidality and bipolar-related symptoms, respectively. Given the significance of clinical understanding, two psychiatrists validate the annotated dataset with a pairwise annotator agreement of 0.77 and a group-wise agreement of 0.88. Unlike the existing datasets~\cite{gaur2019knowledge,shing2018expert,jagfeld2021understanding,vsnajder2018not}, the proposed dataset both includes (i) future suicidality of BD patients and (ii) a user's mood history that can be important features for diagnosing mood episodes~\cite{ortiz2018episode} and future suicidality~\cite{johnson2017suicidality}. 

Based on the developed BD dataset, we propose a novel multi-task learning framework to jointly learn (i) the future suicidality of a given BD user and (ii) their BD symptoms over time. Since a BD symptom can contribute to future suicidality differently depending on when it occurs, we suggest a temporal symptom-aware attention method to determine which symptoms are the most influential for predicting future suicidality over time. 
In particular, the proposed multi-task learning model has three components: (i) the contextualized post-encoder, (ii) a temporal symptom-aware attention layer, and (iii) a task-dependent multi-task decoder. 
After the model generates post representations using the pre-trained Sentence-BERT (SBERT)~\cite{reimers2019sentence} in the contextualized post encoder, the bi-LSTM layer encodes a sequential context of post representations considering variable time intervals between posts. The temporal symptom-aware attention layer then calculates the attention weights of posts to give more weight to critical symptoms affecting the risk classification decision. Finally, the multi-task decoder estimates the probability of future suicidality levels and BD symptoms. For effective multi-task learning, we sum up the losses for each task using the uncertainty weight loss method~\cite{kendall2018multi}, evaluating the task-dependent uncertainty of each task. The proposed model can capture the progressive patterns of BD symptoms and outperform the state-of-the-art methods for predicting future suicidality by leveraging the benefits of multi-task learning. Furthermore, investigating the attention weights based on BD symptoms helps to interpret how they affect the user's future suicidality over time. The provided interpretability from our model can support clinicians in improving their understanding of connections between psychiatric conditions and allowing proper interventions for at-risk people.

We summarize the contributions of this work as follows.
\begin{itemize}[leftmargin=3mm]
    \item We release our codes and a novel BD dataset\footnote{\url{https://sites.google.com/view/daeun-lee/dataset/kdd-2023}}, which contains both the future suicidality and BD symptoms labels, validated by two psychiatrists. The dataset can benefit researchers aiming to develop methods for suicide prevention.
    
    \item To the best of our knowledge, this is the first study that proposes a multi-task learning model for predicting the future suicidality of BD patients on social media by leveraging the knowledge of bipolar symptoms (i.e., manic mood, somatic complaints). The model can accurately capture bipolar symptom transition patterns and outperform the state-of-the-art methods for detecting future suicidality.
    
    \item The proposed temporal symptom-aware attention method provides interpretability, which can help clinicians understand BD patients more comprehensively, thereby providing timely interventions by tracking mental state progression.
    %We propose ~\delee{a deep encoder-decoder, multi-task sequence model} to forecast the future suicidal risk of bipolar disorder by monitoring users' historical trajectories of BD symptoms. Our evaluation results on the real-world dataset reveal that the proposed model i) captures the transition pattern from bipolar to suicidal ideation and ii) outperforms the state-of-the-art methods for detecting future suicide ideation.    
    % \item Our analysis results demonstrate that the diverse factors affecting users' future suicide risk on social media datasets can be compared to a clinical trial, which helps understand the lived experience of bipolar disorder.
\end{itemize}

\section{Related Work} 

\textbf{Social Media to Understand Bipolar Disorder.} With the proliferation of social media, many studies have attempted to address the severe social problems of BD using user activity data on social media~\cite{sawhney2021towards,shing2018expert,kim2020deep}. For example, \citet{vsnajder2018not} showed differences in language use between users with and without BD on Reddit, and this characteristic has been utilized to discover the risk of BD~\cite{coppersmith2014quantifying, cohan2018smhd, vsnajder2018not}. Several studies have analyzed (i) the living experience of BD patients~\cite{jagfeld2021understanding, mandla2017being, sahota2020bipolar} and (ii) an understanding of how people perceive their mental states and share their experiences~\cite{yoo2019semantic} using social media data through qualitative studies. However, while the prior work on BD analysis has revealed valuable insight into the characteristics of individuals with BD, little attention has been paid to predicting the future suicidality of BD patients. Because suicidal ideation can often be developed into an actual attempt~\cite{bryan2016importance,bryan2018predictors,klonsky2016suicide,nock2008cross}, such a model that predicts future suicidality of BD patients can be used for BD patients who usually have suicidal ideation~\cite{harvey2022natural}.

%However, bipolar disorder studies using social media data are still insufficient compared to the severity of the disease than others ~\cite{kim2021machine}. 
%Also, the datasets for considering bipolar in social media are not disclosed or do not contain future suicidality and symptom information~\cite{vsnajder2018not,jagfeld2021understanding}, one of the most critical factors in predicting suicidal tendencies. 
%To this end, we created an experimental dataset labeled future suicidal risk and bipolar symptoms. This dataset is validated by psychiatrists and publicly available. We believe the dataset can be helpful for researchers who want to assess suicidal ideation using social media data for early intervention for people at suicidal risk.
   
\noindent  
\textbf{Future Suicidality Assessment Using Social Media Data.} 
While most of the work has focused on identifying the current suicidality revealed in a given post from social media~\cite{lee2022detecting, sawhney2021phase, lee2020cross, sawhney2020time, azim2022detecting}, a few studies have investigated monitoring a transition of users who have not yet shown suicidality but would potentially reveal it in the future. \citet{lekkas2021predicting} strived to predict whether adolescents will show suicidal intentions within a month of using Instagram with an ensemble model. Similarly, \citet{de2016discovering} attempted to discover the current suicidality of individuals who posted on mental-health-related communities on Reddit by identifying whether a user would write on SuicideWatch, a suicide-related subreddit. They found that users who show suicidality tend to reveal changes in linguistic structures, interpersonal awareness, and social interactions. Unlike the previous work, we predict future suicidality of BD patients by considering the past temporal transition behavior since BD patients suffer from such mood change symptoms. To the best of our knowledge, this is the first work that proposes a future suicidality prediction model by jointly learning two tasks, predicting (i) future suicidality and (ii) BD symptoms.

\section{Bipolar Disorder Data} ~\label{sec:data} \\ [-0.3in]
\subsection{Data Collection and Preprocessing}
    \textbf{Collecting Data.} We collected posts published between January 1st, 2008, and September 30th, 2021, from the three representative bipolar-related subreddits, including \textit{r/bipolar} (BPL), \textit{r/BipolarReddit} (BPR), and \textit{r/BipolarSOs} by using the open-source \textit{Reddit} API\footnote{\url{https://www.reddit.com/dev/api/}}.
    To identify individuals who exhibit suicide ideation, we also collected all the posts during the same data collection period from \textit{r/SuicideWatch,} where people share their suicidal thoughts with others.
    Among the collected posts, we used the posts written by users who have been professionally diagnosed with BD~\cite{jagfeld2021understanding} in this study. For example, users who reported BD diagnosis, e.g., a user who wrote, ``I was diagnosed with Bipolar type-I last year.'', were included in our study. Thus, posts regarding BD symptoms of other individuals, including family members or friends but not themselves, were excluded. Finally, our dataset contains 7,592 posts published by 818 users, i.e., BD patients.
    %We used the open-source \textit{Reddit} API~\footnote{\url{https://www.reddit.com/dev/api/}} for collecting posts from bipolar-related subreddits; bipolar (r/bipolar (BPL), r/BipolarReddit (BPR), and r/BipolarSOs), which were published between January 1, 2008, and September 31, 2021. During the same period, we also collected all posts from r/SuicideWatch to find users who showed suicide ideation. However, we do not characterize differences between the posts from bipolar-related subreddits and r/SuicideWatch during the annotation process because we observed that users with bipolar tend to show their suicidal ideation in both communities. Through this process, we collect ~\delee{0000 posts and 0000} users and 000 posts.
    %     We then connected the posts in the bipolar-related subreddits to the posts in \textit{r/SuicideWatch} since users with BD tend to disclose their suicidal thoughts in both communities. 

    \noindent
    \textbf{Preprocessing Data.} 
    We first anonymized the collected posts by removing information that could be used as personal identifiers. We then converted the texts to lowercase, removed special characters, striping whitespaces, and stopwords, and lemmatized them. %After preprocessing, \delee{0000 posts} are left in the dataset.
    %We anonymize the collected posts by removing personally identifiable information and all named entities in the text. We then convert the text to lowercase, remove special characters, striping whitespaces, and stopwords, and lastly lemmatize the text. After preprocessing, ~\delee{0000 posts.} are left.

\begin{table}[]
\centering
\caption{Summary of annotated labels in our BD data.}
\label{tab:annot_result}
\resizebox{0.95\linewidth}{!}{%
\begin{tabular}{l|c|lr}
\hline
 & Total Num & Category     & Num (\%)   \\ \hline \hline
\multirow{3}{*}{\begin{tabular}[c]{@{}l@{}}A. Diagnosed \\ Bipolar Disorder Type\end{tabular}} & \multirow{3}{*}{818 users}   & BD-I      & 224 (27.3\%)  \\
 &           & BD-II        & 501 (61.2\%)  \\
 &           & NOS          & 93  (11.3\%)  \\ \hline
\multirow{9}{*}{\begin{tabular}[c]{@{}l@{}}B. Bipolar Disorder \\ Symptom\end{tabular}}        & \multirow{6}{*}{7,592 posts} & Depressed & 3,628 (47.7\%)\\
 &           & Manic        & 981  (12.9\%)\\
 &           & Anxiety      & 859  (11.3\%) \\
 &           & Irritability     & 508  (6.6\%) \\
 &           & Remission & 523  (6.8\%) \\
 &           & Other        & 1,093 (14.3\%) \\ \cline{2-4} 
                                                                                               & \multirow{3}{*}{1,747 posts} & Somatic   & 1,293 (74.0\%) \\
 &           & Psychosis    & 429 (24.5\%) \\
 &           & Both         & 25 (1.4\%)   \\ \hline
\multirow{4}{*}{C. Suicidality}                                                               & \multirow{4}{*}{7,592 posts} & Indicator & 6,302 (83.0\%)\\
 &           & Ideation     & 918 (12.0\%)  \\
 &           & Behavior     & 266 (3.5\%)  \\
 &           & Attempt      & 106 (1.3\%)  \\ \hline 
\end{tabular}%
}
\vspace{-0.17in}
\end{table}

\subsection{Annotation Process}
    To label the collected Reddit dataset, we recruited four researchers, knowledgeable in psychology and fluent in English, as annotators. With the supervision of a psychiatrist, the four trained annotators labeled 818 users and their 7,592 anonymized Reddit posts using the open-source text annotation tool \textit{Doccano}~\cite{doccano}. During annotations, we mainly consider two different label categories: (i) BD symptoms (e.g., manic, anxiety) and (ii) suicidality levels (e.g., ideation, attempt). We further annotate the diagnosed BD type (e.g., BD-I, BD-II) for data analysis. If there is any conflict in the annotated labels across the annotators, all the annotators discuss and reach an agreement under the supervision of the psychiatrists. The information about the final annotated labels in our data is summarized in Table~\ref{tab:annot_result}. We now briefly describe the definition of each category; more details about each category and the corresponding examples are described in Appendix~\ref{sec:appendix_annot_guide}.

    \textbf{A. Diagnosed Bipolar Disorder Types:} 
        We label users into one of the three BD diagnosis types based on the self-report in their posts. The BD diagnosis types include \textit{Bipolar Disorder-I} (BD-I), \textit{Bipolar Disorder-II} (BD-II), and \textit{Not Otherwise Specified Bipolar Disorder} (NOS) based on the Diagnostic and Statistical Manual of Mental Disorders (DSM-5)~\cite{APA2013} and the International Statistical Classification of Diseases and Related Health Problems (ICD-10)~\cite{world2016international}. 
    \textbf{B. Bipolar Disorder Symptoms:}
        We employed the Bipolar Inventory of Symptoms scale (BISS)~\cite{bowden2007development} to cover mood polarity (i.e., manic, depressed) and the spectrum of BD symptomatology (i.e., psychosis, somatic complaints). Accordingly, we first annotate mood symptoms consisting of \textit{Depressed}, \textit{Manic}, \textit{Anxiety}, \textit{Remission}, \textit{Irritability}, and \textit{Other}. Note that \textit{Other} covers moods that do not fall into the other five mood symptoms. If we find any BD somatic symptoms, we annotate an additional somatic symptom label that includes \textit{Somatic complaint}, \textit{Psychosis}, and \textit{Both}. 
    \textbf{C. Levels of Suicidality:}
        We also annotate the posts to categorize them into different suicidality levels. We utilize the existing criteria from \cite{shing2018expert, gaur2019knowledge} that provide common five levels of suicidality, including \textit{No risk}~(NR), \textit{Suicide Indicator}~(IN), \textit{Suicidal Ideation}~(ID), \textit{Suicidal Behavior}~(BR), and \textit{Actual Attempt}~(AT), based on the Columbia Suicide Severity Rating Scale (C-SSRS)~\cite{posner2011columbia}. We merge \textit{No Risk} with \textit{Suicide Indicator} since people with bipolar disorder are already considered more at risk than the general population in suicide~\cite{rihmer2002bipolar,ilgen2010psychiatric}.

\begin{table}[]
\centering
\caption{Expert validation results in our BD data. (E1, E2: Expert Psychiatrists / I: Internal Annotators)}
\label{tab:cohen}
\vspace{-0.1in}
\resizebox{\columnwidth}{!}{%
\begin{tabular}{c|ccc|ccc|ccc}
\hline
                        & \multicolumn{3}{c|}{\textbf{Suicidality}} & \multicolumn{3}{c|}{\begin{tabular}[c]{@{}c@{}}\textbf{Mood}\\ \textbf{Symptom}\end{tabular}} & \multicolumn{3}{c}{\begin{tabular}[c]{@{}c@{}}\textbf{Somatic} \\ \textbf{Symptom}\end{tabular}} \\ \hline \hline
\textbf{Krippendorff's $\alpha$} & \multicolumn{3}{c|}{0.88}        & \multicolumn{3}{c|}{0.76}                                                   & \multicolumn{3}{c}{0.72}                                                       \\ \hline \hline
\textbf{Cohen's $\kappa$}        & E1     & E2      & I    & E1                       & E2                       & I                     & E1                        & E2                        & I                      \\ \hline
E1                      & 1               & -       & -    & 1                        & -                        & -                     & 1                         & -                         & -                      \\
E2                      & 0.89            & 1       & -    & 0.81                     & 1                        & -                     & 0.80                      & 1                         & -                      \\
I                       & 0.77            & 0.77    & 1    & 0.72                     & 0.66                     & 1                     & 0.77                      & 0.66                      & 1                      \\ \hline
\end{tabular}%
}
\end{table}

%\multirow{2}{*}{\textbf{Cohen's $\kappa$}} & \multicolumn{3}{c}{\textbf{Suicidal Risk}}& \multicolumn{3}{c}{\textbf{Mood Symptom}}& \multicolumn{3}{c}{\textbf{Somatic Symptom}} \\ 
%\cline{2-10}& \textbf{E1} & \textbf{E2} & \multicolumn{1}{c|}{\textbf{I}} & \textbf{E1} & \textbf{E2} & \multicolumn{1}{c|}{\textbf{I}} & \textbf{E1}& \textbf{E2}& \textbf{I}\\ \hline 
\subsection{Evaluation of Annotation}
    \subsubsection{Expert Validation.} Since the accuracy of the labels (e.g., suicidality, BD symptoms) in the dataset is crucial, we validate the annotated BD dataset with two psychiatrists, as domain experts, by providing 212 randomly selected posts published by 25 users. Table~\ref{tab:cohen} summarizes the Krippendorff’s alpha-reliability~\cite{krippendorff2018content} and Cohen’s Inter-Annotator Agreement~\cite{cohen1960coefficient} among the experts and annotators. The results suggest that our annotations in the dataset are reliable as the overall Krippendorff scores show high agreement, for example, 0.88 for suicidality and 0.76 for mood symptoms, which is similar or even higher than previous studies (e.g., $\alpha$=0.69~\cite{gaur2019knowledge}). The maximum and minimum pairwise Cohen's scores present a fair agreement of 0.89 and 0.66, respectively. 

\begin{table}[]
\caption{Comparisons with existing datasets.}
\label{tab:data-comp}
\vspace{-0.1in}
\resizebox{\linewidth}{!}{
\begin{tabular}{lccccc}
\hline
                                                                               & \multirow{2}{*}{\textbf{Ours}} & \multicolumn{2}{c}{\textbf{\begin{tabular}[c]{@{}c@{}}Suicidality\\ datasets\end{tabular}}} & \multicolumn{2}{c}{\textbf{\begin{tabular}[c]{@{}c@{}}Bipolar Disorder\\ datasets\end{tabular}}} \\ \cline{3-6} 
                                                                               &                                & \textbf{\begin{tabular}[c]{@{}c@{}}Gaur et al\\ \cite{gaur2019knowledge}\end{tabular}}                           & \textbf{\begin{tabular}[c]{@{}c@{}}Shing et al\\ \cite{shing2018expert}\end{tabular}}                          & \textbf{\begin{tabular}[c]{@{}c@{}}Jagfeld et al\\ \cite{jagfeld2021understanding}\end{tabular}}                          & \textbf{\begin{tabular}[c]{@{}c@{}}Sekulić et al\\ \cite{sekulic2018not}\end{tabular}}                         \\ \hline \hline
\textbf{\begin{tabular}[c]{@{}l@{}}Current \\ Suicidality\end{tabular}}        & \cmark                         & \cmark                                        & \cmark                                        & \xmark                                          & \xmark                                         \\ \hline
\textbf{\begin{tabular}[c]{@{}l@{}}Future \\ Suicidality\end{tabular}}         & \cmark                         & \xmark                                        & \xmark                                        & \xmark                                          & \xmark                                         \\ \hline
\textbf{BD Diagnosis}                                                          & \cmark                         & \xmark                                        & \xmark                                        & \cmark                                          & \cmark                                         \\ \hline
\textbf{BD Symptom}                                                            & \cmark                         & \xmark                                        & \xmark                                        & \xmark                                          & \xmark                                         \\ \hline
\textbf{\begin{tabular}[c]{@{}l@{}}Publicly \\ Available\end{tabular}} & \cmark                         & \cmark                                        & \cmark                                        & \xmark                                          & \xmark                                         \\ \hline
\textbf{\begin{tabular}[c]{@{}l@{}}Expert \\ Validation\end{tabular}}          & \cmark                         & \cmark                                        & \cmark                                        & \xmark                                          & \xmark                                         \\ \hline
\textbf{Duration}                                                              & 2008-2021                      & 2005-2016                                     & 2008-2015                                     & 2006-2019                                       & 2005-2018                                      \\ \hline
\textbf{\# of users}                                                            & 818                            & 500                                           & 934                                           & 19,685                                          & 3,488                                          \\ \hline
\textbf{\# of posts}                                                            & 7,592                          & 15,755                                        & -                                             & 21,407,595                                      & -                                              \\ \hline
\end{tabular}}
\vspace{-0.15in}
\end{table}

\subsubsection{Comparison with Existing Datasets.}
    We compare our BD dataset with four widely-used datasets~\cite{gaur2019knowledge,shing2018expert,jagfeld2021understanding,sekulic2018not} from prior studies on suicide and mental health in Table~\ref{tab:data-comp}. First, we find that only two datasets have been released publicly, while the other datasets have not been disclosed. More importantly, no existing dataset has both suicidality and BD symptom labels. The existing suicide datasets~\cite{gaur2019knowledge,shing2018expert} do not include BD-specific and future suicide information. The prior BD datasets~\cite{jagfeld2021understanding,sekulic2018not} have no suicidality labels, and their BD diagnosis labels were generated computationally without expert validation. The proposed BD dataset, on the other hand, includes future suicidality and BD symptom labels, which are validated by clinical experts.

\subsection{Timeline (Post Sequence) Construction}~\label{sec:window}
\begin{figure}[h]
    \centering
    \includegraphics[width=.95\linewidth]{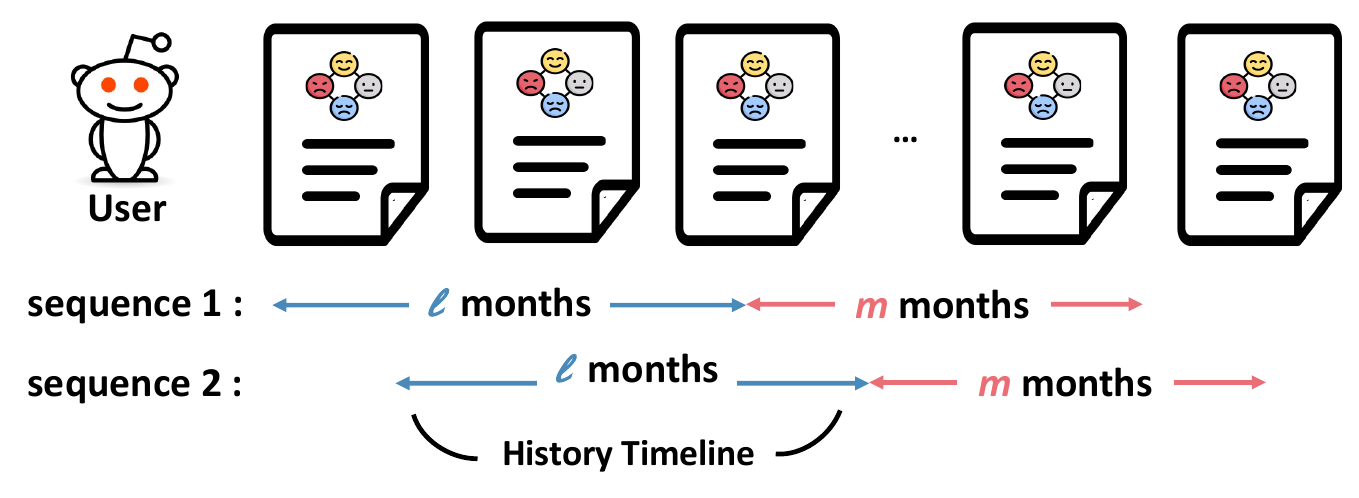}
    \caption{Post sequences (timelines) construction by considering temporal sliding windows.}~\label{fig:days}
    \vspace{-0.2in}
\end{figure}

\noindent
    BD patients tend to have more severe symptom changes over time than other mental disorders; thus, suicidality constantly changes. Since users' posts on different timelines show disparate BD symptoms and suicidality levels, it is crucial to predict future suicidality in each timeline and understand how long past posts should be learned to predict future suicidality. Therefore, we construct multiple timelines (i.e., post sequences) for each user, as shown in Figure~\ref{fig:days}. We set a timeline by selecting the past $l$ months for BD symptom observation and the future $m$ months for suicidality identification. We then slide this timeline window within a given user's posts to obtain multiple post sequences from each user. We assign the future suicidality label as the highest level of the suicidality that appeared in posts over the next $m$ months and exclude sequences with less than three posts in the training period (i.e., past posts). By experimenting with different sets of $l$ and $m$, we set $(l, m) = (6, 1)$, i.e., using the past 6 months for training and the future 1 month for suicidality label extraction, as it shows the best performance. We present the performances of our model with different post-sequence durations in Section~\ref{sec:eq4}. Therefore, we obtain 5,961 post sequences $S$. The distribution of suicidality labels for the sequences is 5,056~(IN), 591~(ID), 215~(BR), and 99~(AT). 
    %Since a user may have multiple timelines, the risk of suicides should be considered for each timeline. Each timeline/window (or each post sequence) includes past $n$ months observations on BD symptoms and future $k$ months observations on suicide risks. By following the sliding window over the multiple post sequences for each user, we acquire successive sequences of $n \in \{ 1,3,6,12\}$ months of past moods and $k \in \{1,3,6,12\}$ months of future suicidal risks for each user as shown in Figure~\ref{fig:days}. Note that we assign the future suicidal risk label for each window as the highest level of the suicide risks that appeared over the next $k$ months among $\{IN, ID, BR, AT\}$. In our work, we set $(n, k) = (6, 1)$, i.e., using the past 6 months of data for training and the future 1 month of data for label extraction, as it shows the best performance; See Section~\ref{sec:eq4} for details. We restrict users who post with at least three posts. As a result, we obtain ~\delee{000} sequences.

\subsection{Data Analysis}
In this section, we analyze our BD dataset to understand distinct BD symptom patterns for BD patients who potentially have high suicidality in the future. We then assess the survival probability using the Kaplan-Meier estimation~\cite{kaplan1958nonparametric} for each BD type (i.e., BD-I, BD-II, and NOS).
%In this section, we analyze the BD dataset to investigate whether there is a distinct BD symptom pattern for BD patients who show future suicide risks. We then assess the survival probability using the Kaplan-Meier estimation~\cite{kaplan1958nonparametric} for each BD subtype (e.g., BD-I, BD-II, and NOS).
% we identify suicidal risk factors in bipolar patients in our dataset based on previous clinical studies. The characteristics demonstrated through the analysis are reflected in a model that predicts the future suicide tendency of bipolar patients to provide more accurate results.
    % figure가 필요할까요? 
    %As illustrated in Figure ~\delee{1}, assume that a user 
    % our goal of predicting whether a user will reveal suicide ideation in the future, we restrict users with at least three posts on mental-illness-related communities (e.g., r/Anxiety, r/Bipolar, and r/BPD). Of these, we found the subset of users that posted in suicide-related communities (e.g., r/SuicideWatch) within the next 12 months. From these users, we apply a window sliding technique by extracting all of the user’s posts. We select one post, which becomes the standard point. Based on that post, we control timeline-length by fixing the timeline (6 months for the front and 12 months for the rear). By repeating this process, we got several data from one user. Finally, to ensure reliability in our dataset, we keep users from overlapping when splitting users into a training set and a test set.
    % 데이터셋 만들기 
    % 1. time sliding 으로 바꾸고 
    % 2. 뒤에 육개월에서 젤 높은 걸 future sucidie risk level 로 봄

\subsubsection{BD Symptoms Affecting Future Suicidality}
    To verify the factors associated with the risk of suicide in the future, we classify the dataset into two groups: i) low-risk group (i.e., IN) and ii) severe-risk group (i.e., ID, BR, AT). We then compare the two groups in terms of the LIWC (Linguistic Inquiry and Word Count)~\cite{pennebaker2001linguistic} results of the users' posts and the annotation results using the t-test. 

    As shown in Table~\ref{tab:T-test}, the target group shows a significantly higher level of past suicidality than the control group. This reveals that a history of suicidality is a significant suicide risk factor in BD patients~\cite{kamali2012associations, malhi2018modeling, abreu2009suicidal, ballard2020symptom}. Furthermore, we observe that the severe-risk group shows more elevated depressed mood, irritability, and psychosis than the low-risk group but is less manic. This observation aligns with the clinical studies that identified dominant depression mood~\cite{xue2021suicidality, simpson1999risk, ballard2020symptom,abreu2009suicidal}, irritability~\cite{ballard2020symptom, gonda2012suicidal}, and psychotic features~\cite{miller2020bipolar, gonda2012suicidal} as major suicide risk factors in BD patients, but the mania status is not significantly related~\cite{miller2020bipolar}. We also find similar results in the LIWC categories, which reveal higher values in \textit{negemo}, \textit{sad}, and \textit{anger} for the severe-risk group. Unlike previous studies~\cite{abreu2009suicidal, ballard2020symptom,gonda2012suicidal}, the ratios of anxiety for the two groups are not statistically different. This implies that social media posts make it difficult to detect clinical anxiety accompanied by physical symptoms like agitation, raised blood pressure, or sweating~\cite{kazdin2000encyclopedia}. 
    
    Furthermore, we compare the social characteristics of the two groups. We discover that the target group uses more family-related words. It could link to negative experiences with family, which often affect their lives, such as a lack of family support, divorce, or unmarried~\cite{dome2019suicide, gonda2012suicidal}. We also find that most people who mentioned work-related words belong to the control group, indicating they might be paid employees or students. According to the previous study~\cite{dome2019suicide, gonda2012suicidal}, unemployment is also associated with higher suicide rates. Overall, the analysis results demonstrate that the diverse symptom-related factors affecting users' future suicidalities revealed in social media data show a similar pattern with a clinical trial, which helps understand the living experience of BD patients when clinicians make decisions. More details are included in Table~\ref{tab:total T-test} in Appendix.

% \begin{table}[t]
% \centering
% \caption{Differences between the target (severe-risk) and control (low-risk) groups. * indicates the p-value of the feature is less than 0.05, which is considered highly statistically significant.}
% \label{tab:T-test}
% \vspace{-0.1in}
% \resizebox{\linewidth}{!}{%
% \begin{tabular}{@{}cc|cc|cc@{}}
% \toprule
% \textbf{Suicide History} & \textbf{t(p)}          & \textbf{BD Symptom}   & \textbf{t(p)}         & \textbf{LIWC}    & \textbf{t(p)}         \\ \hline \hline
% Indicator              & -11.91(0.00*) & Depressed    & 5.43(0.00*)  & negemo  & 3.51(0.00*)  \\
% Ideation        & 9.50(0.00*)   & Manic        & -3.88(0.00*) & anger   & 2.85(0.00*)  \\
% Behavior        & 6.59(0.00*)   & Irritability & -6.07(0.00*) & sad     & 5.16(0.00*)  \\
% Attempt         & 3.39(0.00*)   & Anxiety      & -0.35(0.72)  & death   & 8.03(0.00*)  \\
%                 &               & Remission    & -0.28(0.77)  & family  & 2.23(0.03*)  \\
%                 &               & Somatic      & 1.11(0.267)  & work    & -2.38(0.02*) \\
%                 &               & Psychosis    & 2.44(0.01*)  & achieve & -2.59(0.01*) \\ \bottomrule
% \end{tabular}%
% }
% \vspace{-0.15in}
% \end{table}

% \usepackage{graphicx}
\begin{table}[]
\centering
\caption{Differences between the target (severe-risk) and control (low-risk) groups. * indicates the p-value of the feature is less than 0.05 (**: p<0.005), which is considered highly statistically significant.}
\label{tab:T-test}
\vspace{-0.1in}
\resizebox{\columnwidth}{!}{%
\begin{tabular}{lc|cc|cc}
\hline
\textbf{Suicidality} & \textbf{t}    & \begin{tabular}[c]{@{}c@{}}\textbf{BD}\\ \textbf{Symptom}\end{tabular} & \textbf{t}   & \textbf{LIWC}    & \textbf{t}  \\ \hline \hline
Indicator   & -11.91** & Depressed                                            & 5.43**  & negemo  & 3.51** \\
Ideation    & 9.50**   & Manic                                                & -3.88** & anger   & 2.85** \\
Behavior    & 6.59**   & Irritability                                         & -6.07** & sad     & 5.16** \\
Attempt     & 3.39**   & Anxiety                                              & -0.35   & death   & 8.03** \\
            &          & Remission                                            & -0.28   & family  & 2.23*  \\
            &          & Somatic                                              & 1.11    & work    & -2.38* \\
            &          & Psychosis                                            & 2.44*   & achieve & -2.59* \\ \hline
\end{tabular}%
}
\vspace{-0.15in}
\end{table}

\subsubsection{Survival Analysis}
\begin{figure}[h]
    \centering
    \includegraphics[width=0.90\linewidth]{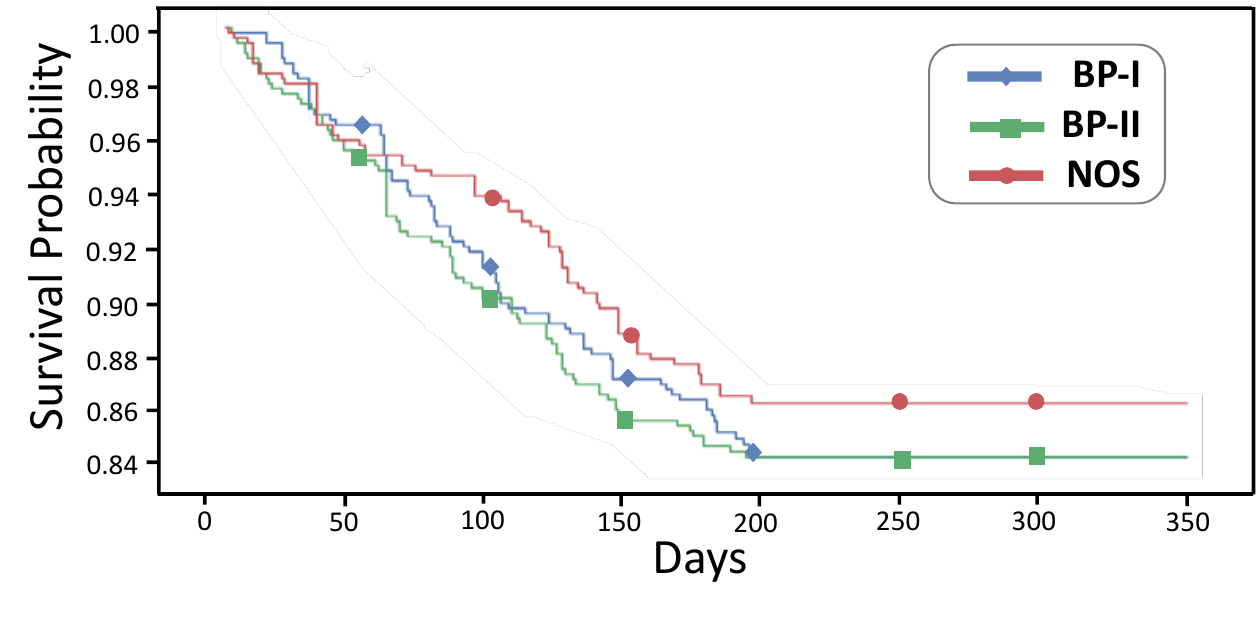}
    \vspace{-0.15in}
    \caption{Analysis of survival probability for BD types.}
    \label{fig:SurvivalAnalysis}
    \vspace{-0.2in}
\end{figure}

We next assess the survival probability for each BD subtype (i.e., BD-I, BD-II, and NOS) using the Kaplan-Meier estimation~\cite{kaplan1958nonparametric}. Following the estimation method \cite{kaplan1958nonparametric}, we observe 180 days after a certain time to verify whether a user is still alive in our dataset. Note that we assume a user has not survived if the user has never posted within the observation period. Figure~\ref{fig:SurvivalAnalysis} shows that BD-II patients have the lowest survival rate, followed by BD-I. This interpretation aligns with the prior clinical studies that present BD-II as having higher suicidality than BD-I, and the rapid cycling of BD-II is hazardous~\cite{plans2019completed}.

\subsection{Ethical Concerns} 
We carefully consider potential ethical issues in this work: (i) protecting users' privacies on Reddit and (ii) avoiding potentially harmful uses of the proposed dataset. The Reddit privacy policy explicitly authorizes third parties to copy user content through the Reddit API. We follow the widely-accepted social media research ethics policies that allow researchers to utilize user data without explicit consent if anonymity is protected~\cite{benton2017ethical,williams2017towards}. Any metadata that could be used to specify the author was not collected. In addition, all content is manually scanned to remove personally identifiable information and mask all the named entities. More importantly, the BD dataset will be shared only with other researchers who have agreed to the ethical use of the dataset. This study was reviewed and approved by the Institutional Review Board ((SKKU2022-11-038)).
\section{Future Suicidality Prediction Model for Bipolar Disorder Patients}
\subsection{Problem Statement}
% Given a set of post sequences $\mathbb{P}=\left\{P_{1},P_{2},\cdots,P_{i}\right\}$,

% \begin{table}[]
% \centering
% \caption{ss}
% \label{tab:my-table}
% \resizebox{\columnwidth}{!}{%
% \begin{tabular}{@{}llll@{}}
% symbol & decription              & symbol & description    \\
% l      & 과거 month                & e      & post embedding \\
% m      & 미래 month                & h      & lstm 나온 후      \\
% s      & post sequence(timeline) & g      & att 나온 후       \\
% p      & sequence에 들어있는 post     & a      & att weight     \\
% t      & 포스트 작성 시간               &        &                \\
% n      & post 수                  &        &                \\
% i      & sequence 수              &        &                \\
%       &                         &        &                \\
%       &                         &        &               
% \end{tabular}%
% }
% \end{table}
% has a set of posts $P_{i} = \left\{ p_{t_{1}}^{i},p_{t_{2}}^{i} ,...,p_{t_{n}}^{i} \right\}$ ordered by the posting time where $n$ denotes the number of posts of $s_{i}$ and $t_{n}$ indicates the posting time of the $n_{th}$ post.  can be defined where 

\begin{figure}[t]
    \centering
    \includegraphics[width=1\linewidth]{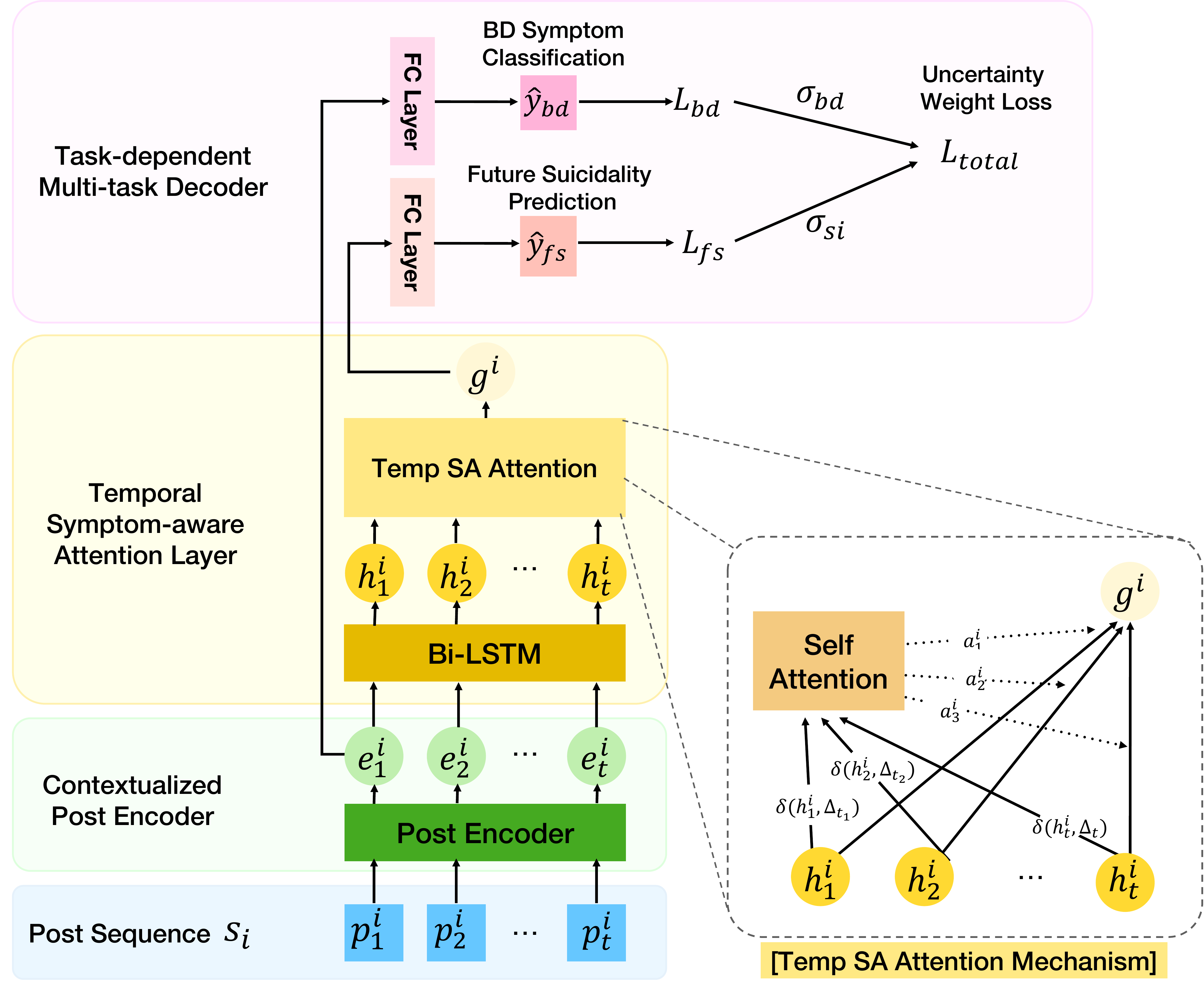}
    \caption{The overall architecture of the proposed multi-task learning model.}~\label{fig:framework}
    \vspace{-0.25in}
\end{figure}

The proposed multi-task learning model aims to (i) predict the future suicidality $y_{fs} \in \left\{IN, ID, BR, AT\right\}$ of $s_{i}$ through a sequence of BD posts $P_{i}$ in the timeline and (ii) classify BD symptoms $y_{bd\_n} \in \left\{\right.$ \textit{No mood, Depressed, Manic, Irritability, Anxiety, and Remission or Somatic complaint and Psychosis}$\left.\right\}$ that appeared in a post $p_{t_{n}}^{i}$. We suppose each post shows one BD mood symptom and, at most, two BD somatic-related symptoms. To be more specific, assume that there is a post sequence $s_{i}\in S=\left\{ s_{1},s_{2},...,s_{i} \right\}$, it can be defined as $s_{i} = \left \{ P^{i},\left\{y_{bd\_n}\right\}_{n=1}^{\left| P^{i} \right|} ,y_{fs} \right \}$. Here, $P_{i} = \left\{ p_{t_{1}}^{i},p_{t_{2}}^{i} ,...,p_{t_{n}}^{i} \right\}$ represents a set of posts ordered by the posting time where $n$ denotes the number of posts of $s_{i}$ and $t_{n}$ indicates the posting time of the $n_{th}$ post. Also, $y_{bd\_n}$ is a set of BD symptom labels of $p_{t_{n}}^{i}$, and $y_{fs}$ is a future suicidality label of $s_{i}$. Note that the time interval between $t_{n}$ and $t_{1}$ is within $l$ months since we take the past $l$ months dataset for feature extraction (See \S \ref{sec:window}). Figure~\ref{fig:framework} illustrates the overall architecture of the proposed model. The model includes three main components: a Contextualized Encoder, a Temporal Symptom-aware Attention Layer, and a Task-dependent Multi-task Decoder.

\subsection{Contextualized Post Encoder} ~\label{sec:model-history}
Each post includes BD-related information about a user. A sequence of posts can show the progressive mood states, which is important information for assessing future suicidality~\cite{johnson2017suicidality}. To generate the semantic representation of each post, we employ the pre-trained Sentence-BERT (SBERT)~\cite{reimers2019sentence}%~\footnote{https://huggingface.co/sentence-transformers/nli-roberta-large}
, which showed promising results in detecting moments of change in the mood~\cite{azim2022detecting} and representing historical tweets ~\cite{sawhney2020time}. SBERT is a modification of the pre-trained BERT network that uses siamese and triplet network structures to derive semantically meaningful sentence embeddings by computing the mean of output vectors for all tokens to derive a fixed-size sentence embedding. We encode each post $p_{t}^{i}$ as follows:
    \begin{equation}
        e_{t}^{i} = SBERT(p_{t}^{i}) \in {\rm I\!R}^{1024}
    \end{equation}
    
\subsection{Temporal Symptom-aware Attention Layer} ~\label{sec:model-att} \\[-0.3in]
    \subsubsection{Sequential context modeling}
    To encode a sequential context of each timeline, we leverage the bidirectional LSTM, a popular method for capturing long-term dependency on social media~\cite{sawhney2021towards,sawhney2021phase,cao2019latent}. Specifically, post-encoding $e_{t}^{i}$ is fed into a Bidirectional LSTM to derive text representation $h_{t}^{i}$. This process is repeated twice, each of which processes the post sequence from left to right (i.e., forward) and right to left (i.e., backward). Finally, the hidden state vectors from each procedure are concatenated as follows:
    \begin{align}
        \overrightarrow{h_{t}^{i}} = LSTM\left ( e_{t}^{i}, \overrightarrow{h^{i}}_{t-1}\right ) \\
        \overleftarrow{h_{t}^{i}} = LSTM\left ( e_{t}^{i}, \overleftarrow{h^{i}}_{t+1}\right ) \\
        h_{t}^{i} = \left [ \overrightarrow{h_{t}^{i}}, \overleftarrow{h_{t}^{i}}\right ]
    \end{align}  

    In this way, the BiLSTM converts the sequence representation of posts $ E  =\left [ e_{t_{1}}^{i}, e_{t_{2}}^{i}, ..., e_{t_{n}}^{i} \right ]$ into contextual representations $H =\left [ h_{t_{1}}^{i}, h_{t_{2}}^{i}, ..., h_{t_{n}}^{i} \right ] \in {\rm I\!R}^{d\times{n}}$ where $d$ is the dimension of the hidden state vector. 
    
    \subsubsection{Temporal symptom-aware attention mechanism}
    We then apply the attention mechanism to pay more attention to a critical mental state affecting the risk classification decision. However, conventional attention mechanisms, such as self-attention ~\cite{vaswani2017attention}, do not consider the BD characteristics that each symptom can contribute differently depending on when it occurs. Since the time intervals between posts may vary considerably, identifying these patterns can be essential in interpreting the mood status over time~\cite{sueki2015association}. Therefore, we propose temporal symptom-aware attention (Temp SA attention) as follows:
    \begin{align}
        g^{i} = \sum_{t=1}^{t_{n}}a_{t}^i h_{t}^i 
        \\
        a_{t}^i = \frac{exp(tanh(\mathcal{F}(\delta(h_{t}^{i}, \Delta_{t}))))}{\sum_{t=1}^{t_{n}}exp(tanh(\mathcal{F}(\delta(h_{t}^{i}, \Delta_{t}))))} \label{eq:selt-att}
        \\
        \delta(h_{t}^{i}, \Delta_{t}) = sigmoid(\theta_{h} -\mu_{h} \Delta_{t})h_{t}^{i} \label{eq:temp-att}
    \end{align}
        where $\mathcal{F}$ is a fully-connected layer and tanh() is the activation function. $\theta_{h}$ is the symptom-specific learnable parameter influenced by $h_{t}^{i}$, and $\mu_{h}$ is also a learnable parameter representing how the influence of $h_{t}^{i}$ changes over time. $\Delta_{t}$ is the time interval between the most recent post $h_{t_{n}}^{i}$ and target post $h_{t}^{i}$. The sigmoid function transforms $\theta_{h} -\mu_{h} \Delta_{t}$ into a probability between 0 and 1. Finally, we derive a sequence representation $g_{i} \in G = \left\{ g_{1},g_{2},...,g_{i} \right\}$ where $a_{t}^{i}$ indicates how symptom-specific information $\delta(h_{t}^{i}, \Delta_{t})$ at $\Delta_{t}$ ago affects the future condition. 
    
\subsection{Task-dependent Multi-task Decoder} ~\label{sec:model-mtl} \\[-0.3in]
    \subsubsection{Future suicidality prediction} 
    To predict the suicidality of each sequence in the future, the proposed decoder generates the final prediction vector as follows: 
   \begin{equation}
        \hat{y}_{fs} = \mathcal{F}_{a}(ReLU(\mathcal{F}_{b}(g_{i})))
        \label{eq: cls}
   \end{equation}
   where $\mathcal{F}_{a},\mathcal{F}_{b}$ are fully-connected layers and $ReLU$ is an activation function.

   Inspired by \citet{sawhney2021towards}, we apply the ordinal regression loss~\cite{diaz2019soft} as an objective function. Rather than employing a one-hot vector representation of the actual labels, a soft encoded vector representation is used to consider the ordering nature between suicidality.
   Assume that $Y_{fs} = \left \{IN = 0, ID = 1, BR = 2, AT = 3  \right \}$ = $\left \{{r_i}_{i=0}^3 \right \}$ denotes ground truth labels, then soft labels are computed as probability distributions $y_{fs}$ = [$y_0, y_1, y_2, y_3$] of $Y_{fs}$ as follows:
   \begin{equation}
       y_{fs\_i} = \frac{e^{-\phi (r_t,r_i)}}{\sum_{k=1}^{\lambda}e^{-\phi (r_t,r_i)}} \forall r_i \in Y_{fs}
   \end{equation}
   where $e^{-\phi (r_t,r_i)}$ is a cost function that penalizes the distance between the actual level $r_t$ and a risk-level $r_i \in Y$, which is formulated as $e^{-\phi (r_t,r_i)} = \alpha \left | r_t-r_i \right |$, where $\alpha$ is a penalty parameter for inaccurate prediction. Finally, the cross-entropy loss is calculated as follows:
   \begin{equation}
       \mathcal{L}_{fs} = - \frac{1}{b}\sum_{j=1}^{b}\sum_{i=1}^{\lambda }y_{fs\_ij}\textrm{log}{\hat{y}_{fs\_ij}}
   \end{equation}
   where $b$ is the batch size, and $\lambda$ is the number of risk levels.

    \subsubsection{BD symptom classification}
    In addition to future suicidality prediction as the main task, we propose to enhance the model by regarding an auxiliary task, bipolar disorder symptom classification. If the post features are good predictors for BD symptoms, the derived information from post features in the auxiliary task can also be leveraged into the main task. By taking the representation, $e_{t}^{i}$ derived from the post encoder layer for each post $p_{t}^{i}$, the model calculates the logits of symptom classification as follows:
   \begin{equation}
        \hat{y}_{bd} = \mathcal{F}_{c}(ReLU(\mathcal{F}_{d}(e_{t}^{i})))
   \end{equation}
     where $\mathcal{F}_{c},\mathcal{F}_{d}$ are fully-connected layers and $ReLU$ is an activation function. BD symptom classification can also be treated as a multi-label classification. Hence, the objective function can be written as follows:
\begin{equation}
       \mathcal{L}_{bd} = - \frac{1}{b}\sum_{j=1}^{b}\sum_{i=1}^{\gamma}y_{bd\_ij}\textrm{log}{\hat{y}_{bd\_ij}} + (1-y_{bd\_ij})\textrm{log}{(1- \hat{y}_{bd\_ij})}
\end{equation}
   where $b$ is the batch size, and $\gamma$ is the number of symptom categories.

    \subsubsection{Multi-task learning} 
    Since our multi-task learning model aims to solve tasks with different scales, i.e., post-level BD symptom prediction and sequence-level suicidality prediction, tuning weights between each task's loss is complicated and costly. Therefore, for effective multi-task learning, we employ the uncertainty weight loss~\cite{kendall2018multi} that weighs multiple loss functions to simultaneously learn various scales of different units by evaluating the task-dependent uncertainty of each task. Finally, the ultimate objective for multi-task learning is summing up the losses.
   \begin{equation}
    \mathcal{L}_{total} = \frac{1}{2\sigma _{fs}^2}\mathcal{L}_{fs}(W) + \frac{1}{2\sigma _{bd}^2}\mathcal{L}_{bd}(W) + log\sigma _{fs}\sigma _{bd}
   \end{equation}
      where $\sigma _{bd},\sigma _{fs}$ are the learnable parameters representing uncertainty for each task, and $W$ is the weight parameter.

\section{Experiments}
% \subsection{Evaluation Metrics}
% To consider the ordinal nature of suicidal risks, we apply the revised metrics of False Positive ($FP$) and False Negative ($FN$) in our experiments as follows~\cite{gaur2019knowledge}.
%     \begin{equation}
%         FP = \frac{\sum_{i=1}^{N_{T}}I(\hat{y_{i}}>y_{i})}{N_{T}}
%     \end{equation}
%     \begin{equation}
%         FN = \frac{\sum_{i=1}^{N_{T}}I(y_{i}>\hat{y_{i}})}{N_{T}}
%     \end{equation}
% where $\hat{y_{i}}$ is the predicted label, $y_{i}$ is the ground truth label for $i^{th}$ test data, and $N_{T}$ is the size of the test data. $\Delta \left ({y_{i}}, \hat{y_{i}}  \right )$ indicates the interval between $y_{i}$ and $\hat{y_{i}}$. The evaluation metrics terms for precision and recall are renamed as graded precision and graded recall, respectively. 

\subsection{Baselines} 
Since predicting future suicidality from BD patients has not been explored in the literature, we compare against baseline approaches from the related tasks, i.e., identifying current risks of suicidality. All the baseline models were developed by considering a sequential context of post representations in detecting suicidality. 
    \begin{itemize}[leftmargin=*]
        % \item  \textbf{SVM+RBF~\cite{amini2016evaluating}}: Each user's embedding vector is fed to a Support Vector Machine with a Radial Basis Function kernel using the kernel parameter $\sigma$ = 0.24 and the cost parameter $c$ = 5.
        % \item \textbf{LR~\cite{de2016discovering}}: The LR is the logistic regression model that uses various hand-crafted features such as linguistic (e.g., a ratio of verbs), interpersonal awareness (e.g., use of the first-person singular), interaction (e.g., number of comments that a user has received) features, and word frequencies.
        % \item  \textbf{MLP~\cite{amini2016evaluating}}: The input vector is provided to an MLP having two hidden layers with 64 dimensions.
        \item \textbf{Suicide Detection Model (SDM)}~\cite{cao2019latent}: The SDM adopts the LSTM layer with an attention mechanism. Fine-tuned FastText embeddings are utilized for encoding posts.
        \item \textbf{C-CNN}~\cite{shing2018expert}: The C-CNN is trained with posts that are encoded by ConceptNet word embeddings ~\cite{speer2017conceptnet}.
        \item \textbf{SISMO}~\cite{sawhney2021towards}: The SISMO uses Longformer~\cite{beltagy2020longformer} and the Bidirectional LSTM to obtain dynamic post embeddings.
        \item \textbf{STATENet}~\cite{sawhney2020time}: The STATENet is a time-aware transformer-based model that uses emotional and temporal contextual cues for suicidality assessment.
        \item \textbf{UoS~\cite{azim2022detecting}}: UoS is the best performing model at the CLPsych 2022 shared task with~\citet{zirikly2019clpsych} to capture moments of change in a suicidal individual's mood. The obtained embeddings from the pretrained Sentence BERT are fed into a biLSTM layer and a multi-head attention layer.
    \end{itemize} 
    
% \sjson{Our work focuses on the multi-task goal that utilizes information on BD symptoms' history to predict suicidality. Yet, none of the existing datasets applied to our multi-task setting, i.e., identifying both BD symptoms and suicidality. So we compared against baseline models that challenged to identify current risks of suicidality (L710). For example, when using [3], [11], [66], and [68], instead of taking the current post as input to identify suicide risk, we used past posts as input to obtain the probability for future suicide risk prediction. In contrast, [65] differed from other baseline models in using two separate encoders for the historical posts and the target post that we wanted to predict suicide risk. Hence, since our dataset consists of only past posts, we only used the encoder that takes the target post as input for the [65] experiment; we also experimented using only the history encoder, but it showed lower performance.
% }

%We categorize the baselines into three approaches: (i) non-sequential modeling and (ii) sequential modeling. Our baselines are summarized as follows.

\subsection{Experimental Settings} 
    To solve the imbalanced data issue, the random oversampling technique~\cite{menardi2014training} is used to generate new train samples by randomly sampling each class independently with the replacement of the currently available samples. All experiments are performed with the stratified 5-fold cross-validation, ensuring that the users in the test set are entirely disjoint and do not overlap with those in the training set. We use 10\% of the training set as validation during training to tune our models' hyper-parameters. For reproducibility, detailed experimental settings are summarized in Appendix~\ref{sec:appendix_exset}.

        % Finally, the dataset includes 1346, 420, 337, 77, and 49 posts for the Supportive, Indicator, Ideation, Behavior, and Attempt levels, respectively. In addition, we implement a stratified 60:20:20 split such that the train, validation, and test sets consist of 1,427, 356, and 446 posts, respectively.
    
% 

\begin{table*}[ht]
\centering
\caption{Performance comparisons of the proposed model and baselines. We report the average of results over 5-fold cross-validation. * indicates that the result is significantly better than
C-CNN (p < 0.05) under Wilcoxon’s Signed Rank test. Bold denotes the best performance and \textit{Italics} denotes the second best.}

\label{tab:eq1}
\vspace{-0.1in}
\resizebox{0.95\textwidth}{!}{%
\begin{tabular}{clccc|ccc|ccc}
\hline
\multirow{2}{*}{Task} & \multicolumn{1}{c}{\multirow{2}{*}{Model}} & \multicolumn{3}{c|}{\textbf{\begin{tabular}[c]{@{}c@{}}4 levels\\ (IN / ID / BR / AT)\end{tabular}}} & \multicolumn{3}{c|}{\textbf{\begin{tabular}[c]{@{}c@{}}3 levels\\ (IN / ID / BR+AT)\end{tabular}}} & \multicolumn{3}{c}{\textbf{\begin{tabular}[c]{@{}c@{}}2 levels\\ (IN / ID+BR+AT)\end{tabular}}} \\ \cline{3-11} 
 & \multicolumn{1}{c}{} & Prec. ↑ & Rec. ↑ & F1 ↑ & Prec. ↑ & Rec. ↑ & F1 ↑ & Prec. ↑ & Rec. ↑ & F1 ↑ \\ \hline
\multirow{8}{*}{Future Suicidality} 
 & STATENet ~\cite{sawhney2020time} & 76.56 & 36.84 & 47.76 & 76.49 & 42.73 & 51.77 & 76.87 & 59.40 & 64.99  \\
 & SISMO ~\cite{sawhney2021towards} & 73.94 & 64.66 & 68.71 & 73.30 & 59.85 & 65.18 & 77.12 & 52.15 & 58.21 \\
 & UoS (STL) ~\cite{azim2022detecting} & 73.14 & 73.11 & \multicolumn{1}{l|}{72.99} & 75.15 & 78.05 & 76.51 & 78.71 & 78.11 & 78.40 \\
 & UoS (MTL All) ~\cite{azim2022detecting} & 73.72 & 75.39 & \multicolumn{1}{l|}{74.47} & 75.33 & 76.39 & 75.82 & 79.01 & 79.30 & 79.15 \\ 
 & SDM ~\cite{cao2019latent} & 72.97 & 73.87 & 73.20 &77.86 & 80.45 & 78.95 & 77.52 & 80.18 & 77.34 \\ 
 & C-CNN~\cite{shing2018expert} & \textit{80.42} & 83.67 & 78.44 & 72.40 & 80.37 & 76.17 & 76.21 & 83.12 & 77.19 \\ \cline{2-11}
 & Ours (STL) & 79.02 & \textit{84.50} & \textit{81.62} & \textit{76.61} & \textit{84.07} & \textit{79.40} & \textit{80.26} & \textit{83.54} & \textit{81.59} \\
 & Ours (MTL All) & \textbf{81.84*} & \textbf{86.58*} & \textbf{82.30*} & \textbf{76.68*} & \textbf{84.47*} & \textbf{79.50*} & \textbf{82.37*} & \textbf{86.52*} & \textbf{82.21*} \\ \hline \hline
\multirow{5}{*}{Bipolar Symptom} & & \multicolumn{3}{c|}{\textbf{8 BD symptoms}} & & & \multicolumn{1}{l|}{} & & & \\ \cline{3-11}
 & UoS (STL) ~\cite{azim2022detecting} & 59.71 & 64.86 & 59.98 & - & - & \multicolumn{1}{l|}{-} & - & - & - \\
 & UoS (MTL All) ~\cite{azim2022detecting} & 57.75 & 70.73 & \multicolumn{1}{l|}{60.66} & - & - & \multicolumn{1}{l|}{-} & - & - & - \\
 & Ours (STL) & 57.62 & 70.65 & \multicolumn{1}{l|}{60.65} & - & - & \multicolumn{1}{l|}{-} & - & - & - \\
 & Ours (MTL All) & \textbf{58.63} & \textbf{70.77} & \textbf{61.24} & - & - & \multicolumn{1}{l|}{-} & - & - & - \\ \hline
\end{tabular}%
}
\end{table*}
%\vspace{-0.1in}

\section{Results}
\subsection{Model Performance}
 % & LR ~\cite{de2016discovering} & 74.95 & 82.14 & 77.16 & 73.36 & 82.09 & 77.06 & 75.19 & 80.60 & 77.42 \\
 % & MLP ~\cite{amini2016evaluating} & 73.21 & 81.65 & 76.63 & 71.86 & 81.09 & 76.13 & 73.73 & 80.66 & 76.72 \\
    Table~\ref{tab:eq1} summarizes the weighted average precision, recall, and F1-score of the proposed model and the baselines for the future suicidality and the bipolar symptom prediction tasks. %Overall, the proposed model outperforms the state-of-the-art models.  
    
        % 3, 2 class 실험 are in majority
    \noindent{\textbf{Future suicidality task}}:
    Since our dataset is disproportionate across the suicide risk levels as shown in Table~\ref{tab:annot_result}, we conduct experiments over the three classification tasks: (i) 2-level, (ii) 3-level, and (iii) 4-level classifications, as shown in Table~\ref{tab:eq1}. For example, we combine AT and BR categories to the highest risk level for the 3-level classification; we merge AT, BR, and ID for the 2-level classification. As shown in Table~\ref{tab:eq1}, we find that the proposed model outperforms all the baseline methods regardless of how the suicide risk level is structured. We observe that STATENet~\cite{sawhney2020time} shows the lowest performance among the baseline methods. That is because STATENet fails to utilize sequential data, while other baselines consider users' posts over time as input sequences. Although other baselines perform better than STATENet by exploiting sequential data, our model surpasses them by learning the temporal dynamics of BD symptoms.
    
    \noindent{\textbf{BD symptom task}}: The results show that our proposed model improves BD symptom prediction performance compared to UoS~\cite{azim2022detecting}. While our model directly utilizes contextualized post embeddings to predict BD symptoms in each post, UoS considers post sequences. This implies that considering the post sequence may interfere with BD symptom prediction of each post since bipolar patients have a characteristic of rapidly changing mood.
    
    \noindent{\textbf{Multi-task learning}}: To evaluate the performance of multi-task learning, we train the proposed model separately for each task (i.e., Single-task learning~(STL)). We find that the proposed multi-task learning (MTL) improves prediction performances from single-task learning (STL) in both BD symptom identification and future suicidality prediction tasks by achieving 61.24\% and 82.30\%, respectively. This suggests that jointly learning BD symptom information helps forecast future risks of suicidality by sharing informative presentations and parameters. 
\begin{table}[]
\centering
\caption{Ablation study results over the proposed model components.}
\label{tab:eq3}
\vspace{-0.1in}
\resizebox{0.90\linewidth}{!}{%
\begin{tabular}{clrr}
\hline
\multicolumn{2}{c}{Model}                      & \multicolumn{1}{l}{Rec.↑} & \multicolumn{1}{l}{F1↑} \\ \hline
\multicolumn{1}{c|}{\multirow{4}{*}{\begin{tabular}[c]{@{}c@{}}Model\\ Component\end{tabular}}} &
  \textbf{Ours (MTL ALL)} &
  \multicolumn{1}{c}{\textbf{86.58}} &
  \multicolumn{1}{c}{\textbf{82.30}} \\
\multicolumn{1}{c|}{} & - w/o Uncertainty Loss & 83.78                     & 80.95                   \\
\multicolumn{1}{c|}{} & - w/o Temp SA Att      & 84.05                     & 81.35                   \\
\multicolumn{1}{c|}{} & - w/o Bi-LSTM          & 87.19                     & 81.69                   \\ \hline
\multicolumn{1}{c|}{\multirow{2}{*}{\begin{tabular}[c]{@{}c@{}}Bipolar\\ Symptom\end{tabular}}} &
  - w/o Somatic &
  86.40 &
  81.77 \\
\multicolumn{1}{c|}{} & - w/o Moods            & 80.27                     & 79.35                   \\ \hline
\end{tabular}%
}\vspace{-0.17in}
\end{table}

\subsection{Ablation Study}
\textbf{Model Component.} We perform an ablation study to examine the effectiveness of each component. Applying the uncertainty weight loss function is a common technique in multi-task learning that can address the challenge of tuning loss coefficients for different tasks with different prediction levels. In our case, the two tasks, suicidality prediction, and BD symptom classification, show different granularities, hence we use the uncertainty parameters to balance their weights during training. This can prevent one task from dominating the objective function and improve the model's overall performance. As shown in Table~\ref{tab:eq3}, there is a significant drop in performance when the uncertainty weight loss is not used.
 Overall, Table~\ref{tab:eq3} shows that the performance is inferior when the self-attention mechanism is applied instead of the proposed temporal symptom-aware attention mechanism. By adding symptom-specific information, we suppose the model can learn that each symptom contributes differently to future suicidality over time. This indicates that not only understanding time intervals but also mood swings over time is essential to predict the future suicidal risk of BD users.

\noindent{\textbf{Bipolar Symptom.}}
We conjecture that the effects of mood and somatic symptom information of BD for predicting future suicidality would differ. To validate this, we train the multi-task model with either mood or somatic symptoms. Note that `w/o somatic' refers to a case where only information on the six mood symptoms is included, but any information on the somatic symptoms is excluded. On the other hand, `w/o mood' refers to a case where only information on the somatic symptoms is included, but any information on the mood symptoms is excluded. As shown in Table~\ref{tab:eq3}, the multi-task model trained with only mood symptoms (`w/o somatic') achieves a higher performance (81.77\% of F1-score) than the model trained with only somatic symptoms (`w/o mood').
Although somatic symptoms are prominent suicidality for BD patients~\cite{miller2020bipolar, gonda2012suicidal}, they appear less frequent than mood symptoms. Thus, the improved performance in the final model (MTL All) signifies that both symptoms play a complementary role in solving the main task.

\subsection{Observational and Predictable Periods} ~\label{sec:eq4}
\begin{figure}[]
\begin{subfigure}{.45\linewidth}
  \centering
  % include first image
  \includegraphics[width=\linewidth]{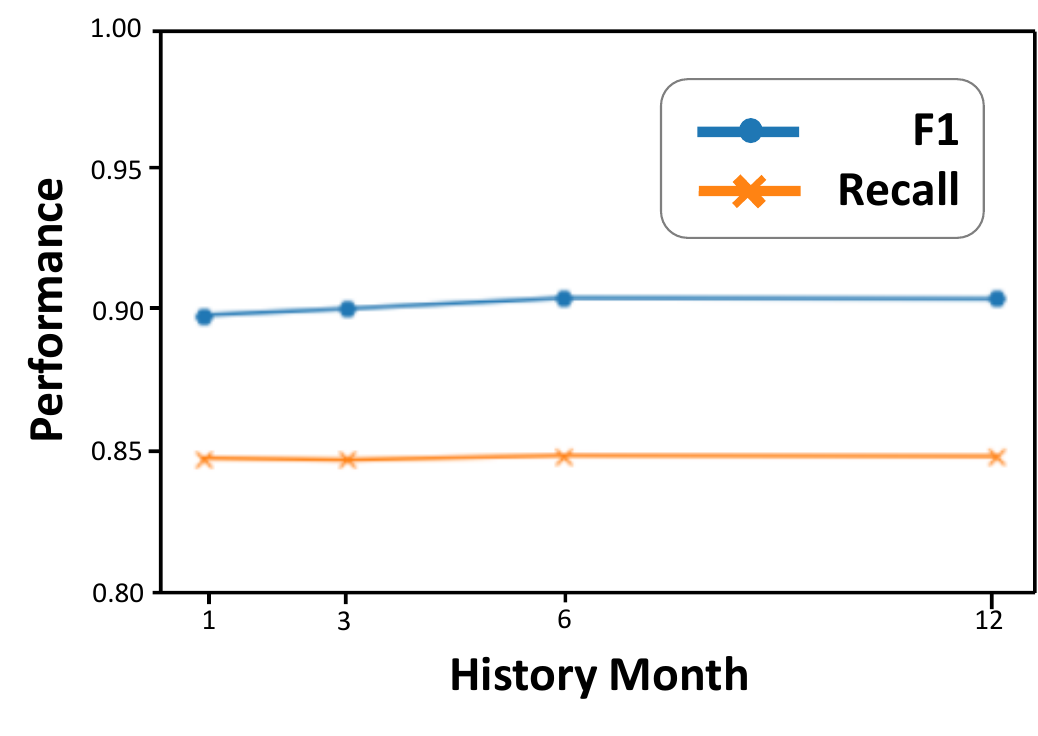}  
  \caption{Varying Observation Period}
  \label{fig:history}
\end{subfigure}
\begin{subfigure}{.45\linewidth}
  \centering
  % include second image
  \includegraphics[width=\linewidth]{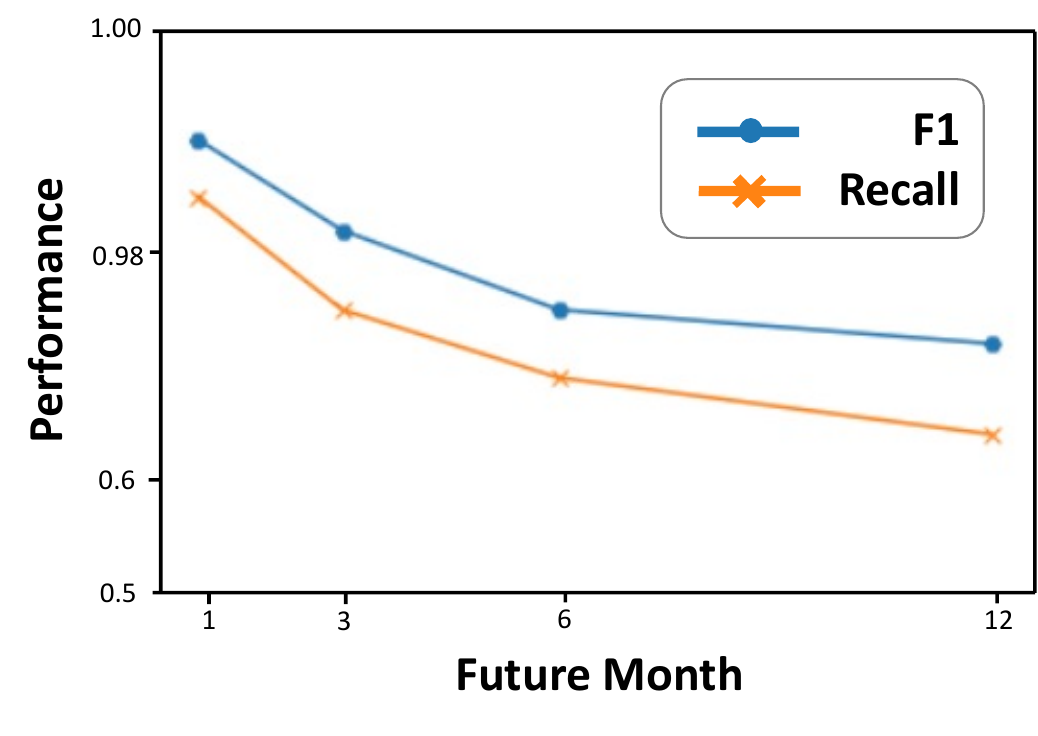}  
  \caption{Varying Forecast Period}
  \label{fig:future}
\end{subfigure}
\vspace{-0.1in}
\caption{Performance of the model by varying observational period $l \in \{ 1,3,6,12\}$ and predictable period $m \in \{1,3,6,12\}$.}
\label{fig:look}
\vspace{-0.2in}
\end{figure}

% 개월 수 비교 해봄
% 보통 다른 곳은 가까이 있는 개월 예측하는게 더 어렵다는데, BD는 달랐음. 이는 무드 변동이 심하기 때문에 과거정보가 너무 먼 미래를 예측하는데는 크게 도움이 안되는 것임. 
As illustrated in Section~\ref{sec:window}, we conduct experiments to find how many months $l \in \{ 1,3,6,12\}$ we should observe in predicting the future suicidality on the next period $m \in \{1,3,6,12\}$. Figure~\ref{fig:history} shows the weighted average F1 score and recall for predicting suicidality in 1 month by training $l \in \{ 1,3,6,12\}$ months. We find that the performance increases as more past days are trained, but no improvement beyond 6 months. This implies that the 12 months observation period is too long to capture informative recent patterns to predict future suicidality. 
However, the longer the future period to be predicted, the worse the performance is trained for 6 months, as shown in Figure~\ref{fig:future}. We interpret that this is because the mood swings of individuals with BD tend to be radical and impulsive~\cite{strejilevich2013mood, o2018mood}, limiting the model's ability to predict the far distant future from historical records. Therefore, the performance of the proposed model is the best when $(l,m) = (6,1)$. According to previous studies, BD patients hospitalized by suicide attempts are likely to commit suicide again between 3 and 6 months after discharge~\cite{desai2005mental}. The proposed model can help offer proper treatment by diagnosing BD symptoms and suicidality early.

\begin{figure*}[h]
    \centering
    \includegraphics[width=1\linewidth]{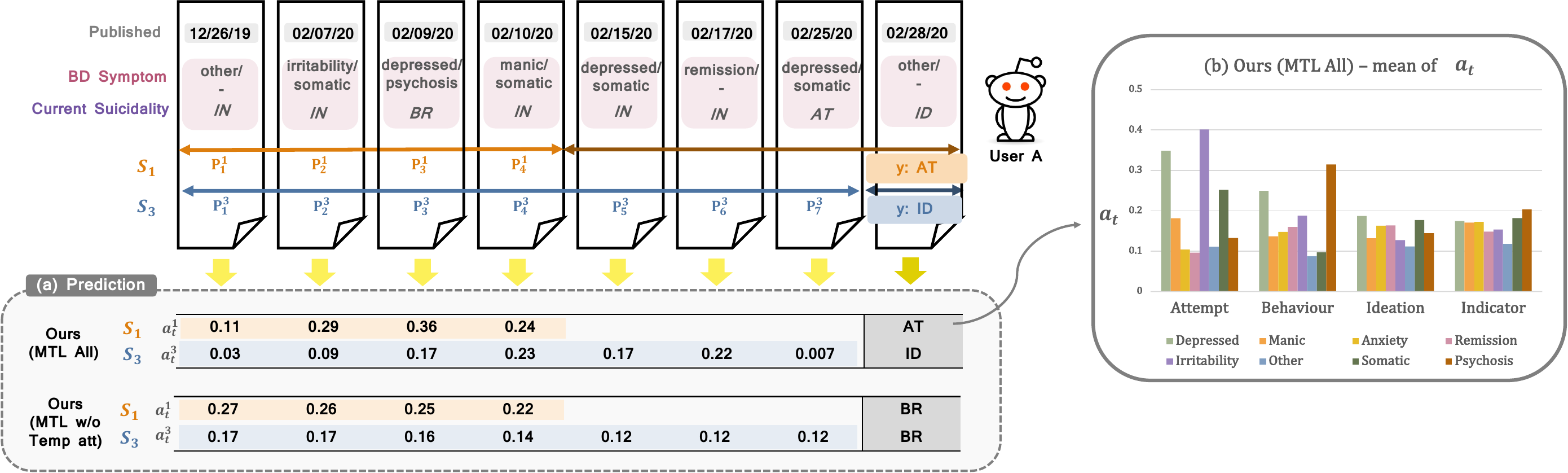}
    \caption{(a) Two example sequences from a single user on how the proposed model assigns attention weights. Such an analysis can provide interpretability in using the proposed model. (b) Average BD Symptoms' temporal attention weights $a_{t}$ depending on the \emph{Future Suicidality} level.}
    \label{fig:qual}
    \vspace{-0.15in}
\end{figure*}

\subsection{Interpretability of the Model}
To demonstrate the interpretability of the proposed model by analyzing attention weights related to BD symptoms, we examine two example sequences, $s_{1}$ and $s_{3}$, extracted from the same user A, where their levels of future suicidality are different. 
In particular, we compare the proposed model with and without the temporal symptom-aware attention mechanism (i.e., `MTL All' vs. `MTL w/o Temp att') in Figure~\ref{fig:qual}.
%We first discuss how whether the temporal symptom-aware attention mechanism is applied affects performance. 
As shown in Figure~\ref{fig:qual}(a), both models correctly predict BD symptoms for each post, but only `MTL all' correctly identifies future suicidality of $s_{1}$ and $s_{3}$. This implies that temporal tracking is useful since bipolar patients have a characteristic of rapid mood swings. We further analyze how the model assigns the temporal symptom-aware attention $a_{t}^{i}$ (in Equation~\ref{eq:selt-att}) to each post over time in predicting future suicidality. In the $s_{1}$ sequence, `MTL All' assigns a lower attention score to $p_{1}^{1}$, whereas giving more attention to $p_{2}^{1}-p_{4}^{1}$. It indicates that \textit{irritability/somatic} and \textit{depressed/psychosis} are crucial for predicting future suicidality~\cite{ballard2020symptom,xue2021suicidality,miller2020bipolar}. Also, we find a similar tendency for the BD symptoms' attention weights in Figure~\ref{fig:qual}(b); as the risk of suicide increases, the importance of attention weights to \textit{anxiety} decreases, while the importance of \textit{depressed} and \textit{irritability} increases significantly.  %이 부분 contribution에 추가해도 될까요

Notably, $s_{3}$ has identical posts with $s_{1}$ (i.e., $p_{1}^{3}-p_{4}^{3}$), but the model differently gives attention weights to them. By focusing on a \textit{manic} symptom in $p_{4}^{3}$, the model forecasts lower suicidality than $s_{1}$, which implies user A's mental status is shown to be improved. Moreover, we find that our model tends to focus on recent posts, giving more attention to the manic symptoms of $p_{4}^{1}$, which is a new observation compared to a previous work that claimed that manic episodes less contribute to future suicidality~\cite{miller2020bipolar}; a future validation is required.
We believe the proposed future suicidality prediction model with an interpretable function, as exemplified in Figure~\ref{fig:qual} can be used for screening and identifying individuals with mental illness on social media to prioritize early intervention for clinical support. 

\section{Concluding Remarks}
In this study, we proposed a novel end-to-end multi-task learning model to jointly learn (i) the future suicidality and (ii) BD symptoms of individuals with BD over time. The proposed model for predicting the future suicidality using temporal symptom-aware attention can (i) accurately capture BD transition patterns and (ii) outperform the state-of-the-art methods for detecting future suicidality for BD patients.
%이전 | By introducing temporal symptom aware attention that can capture which symptoms are the most influential for predicting future suicide, experiment results reveals that the proposed model can (i) accurately capture bipolar symptom transition patterns and (ii) outperforms the state-of-the-art methods for detecting future suicidality for BD patients.
We plan to open our codes and dataset, which contains the future suicidality and BD symptom labels, validated by two psychiatrists. The proposed model and dataset have great utility in identifying the potential suicidality of users in the future, hence preventing individuals from potential suicidality at an early stage.
% We develop a temporal symptom-aware attention mechanism to determine which symptoms are the most influential for predicting future suicide over time.
% predicting future suicidal risk using bipolar symptoms in the extension of sequence modeling for distinguishing current suicidal risk. We propose a temporal symptom-aware attention model to determine which symptoms are the most significant for forewarning future suicide over time. In the contextualized post encoder, the model yields post representations operating the pre-trained Sentence-BERT. After encoding a sequential context of post representations in the bi-LSTM layer regarding uneven time gaps between posts, the temporal symptom-aware attention layer computes the attention weights of posts for giving more attention to a critical symptom involving the risk classification assessment. In the end, the multi-task decoder calculates the possibility of future suicide risk levels and BD symptoms. 

\noindent{\textbf{Clinical Applicability.}}
As an interdisciplinary study, our work contributes to both machine learning and Psychiatry communities. Most BD applications do not provide a suicide warning function and rely solely on self-assessment. On the other hand, we proposed an interpretable and automatic model for predicting the future suicidality of BD patients by introducing a temporal symptom-aware attention mechanism based on sequential context learning.
With the advantage of the proposed model that can help identify complex mood changes and future suicidality in a real context, it can be used for monitoring risks of suicidality for those who are underrepresented in a clinical setting, such as minorities, uninsured people, or patients with a lack of insight. In addition, more concise and timely tracking of mood symptoms can reduce the diagnosis duration, leading to shorter treatment periods and better prognosis in BD patients. Our dataset can help to establish a prevention system for early detection and immediate intervention of BD patients with high-risk suicidality for clinical support. This will enable us to reduce mental health-related social costs and promote public health.

% 안지현 교수님 : 이부분은 해석에 있어서 독립적인 BP 진단 또는 개입 모델인지 오해가 있을 수도 있겠습니다..  우선 임상가의 진단이 가장 일차적이고 이를 기반으로 기능하는 모델로 서술하시면 좋을 것 같습니다.   
% 두서없이 임상가,환자, 연구자의 관점에서 이것저것 써봤는데 내용 develop해서 쓰시면 되지 않을까 싶어요 
% - can help better understanding of complexed mood change in real context
% - a more concise and timely tracking of mood symptoms can lead to/ shorter treatment period and a better prognosis in BD patients/
% - can help to establish a prevention system for early detection and immediate intervention of BD patients with high risk suicidality.  /
% - can monitor suicidal risk for those who to be underrepresented in clinical setting, such as minorities, uninsured people or patients with lack of insight /
% - can reduce mental health related social cost and  promote public health /
% -can develop methodological power and evidence of real-world data in mental health research 

\noindent{\textbf{Limitation.}}
Assessing future suicidality on social media can be subjective~\cite{keilp2012suicidal}, and the analysis of this paper can be interpreted in various ways by the researchers. The experiment data may be sensitive to demographic, annotators, and media-specific biases~\cite{hovy2016social}. Although we carefully selected the users who have been clinically diagnosed as BD based on their Reddit posts~\cite{jagfeld2021understanding}, possibly noisy data can be included if the users misunderstood their diagnoses or did not tell the truth. Moreover, there might be linguistic discrepancies between Reddit and other social media users (e.g., Twitter users) who self-reported BD diagnosis. Lastly, using digital-trace data from social media for predicting mental health can cause low performance depending on the condition of a clinical setting~\cite{de2017language,ernala2019methodological}.

% Nevertheless, an interpretable model can help understand other targets with different statistical patterns and biases~\cite{jacobson2020ethical}.
% Post-traumatic stress disorder may correlate to their demographic features~\cite{preoctiuc2015studying}. Hence, whether these results imply mental health diagnoses or other user characteristics is unclear.

\noindent{\textbf{Future Work.}}
Unfortunately, BD is often misdiagnosed as a depressive disorder since the depressive phase occupies most of the mood episodes~\cite{passos2019machine}. It has been reported that 9 years were taken on average to clinically diagnose BD~\cite{ghaemi2007bipolar}, which can potentially delay treatment opportunities and increase the risk of suicide~\cite{keramatian2022clinical}. Further research could focus more on comparing similar symptoms in different diagnoses to make precise detection (e.g., depressed mood in major depression). We also aim to apply the proposed model to data collected in the clinical field, such as EMR data, to validate the proposed model's effectiveness to determine the potential for practical applications.
\vspace{-0.1in}
\begin{acks}
We would like thank Dr. Ji Hyun An (M.D., Ph.D.) and Myung Hyun Kim (M.D.) for their thoughtful advice on experimental design and clinical data validation. We also thank Chaewon Park for helping data annotation and statistical analysis. This research was supported by the Ministry of Education of the Republic of Korea and the National Research Foundation of Korea (NRF-2022S1A5A8054322), and by the National Research Foundation of Korea (NRF) grant funded by the Korea government (MSIT) (No. 2023R1A2C2007625).
\end{acks}

\bibliographystyle{ACM-Reference-Format}
\balance
\bibliography{reference}

%\clearpage
\clearpage
\appendix
\section{Annotation Criteria} \label{sec:appendix_annot}

 \begin{figure}[htb!]
  \centering
  \includegraphics[width=.95\linewidth]{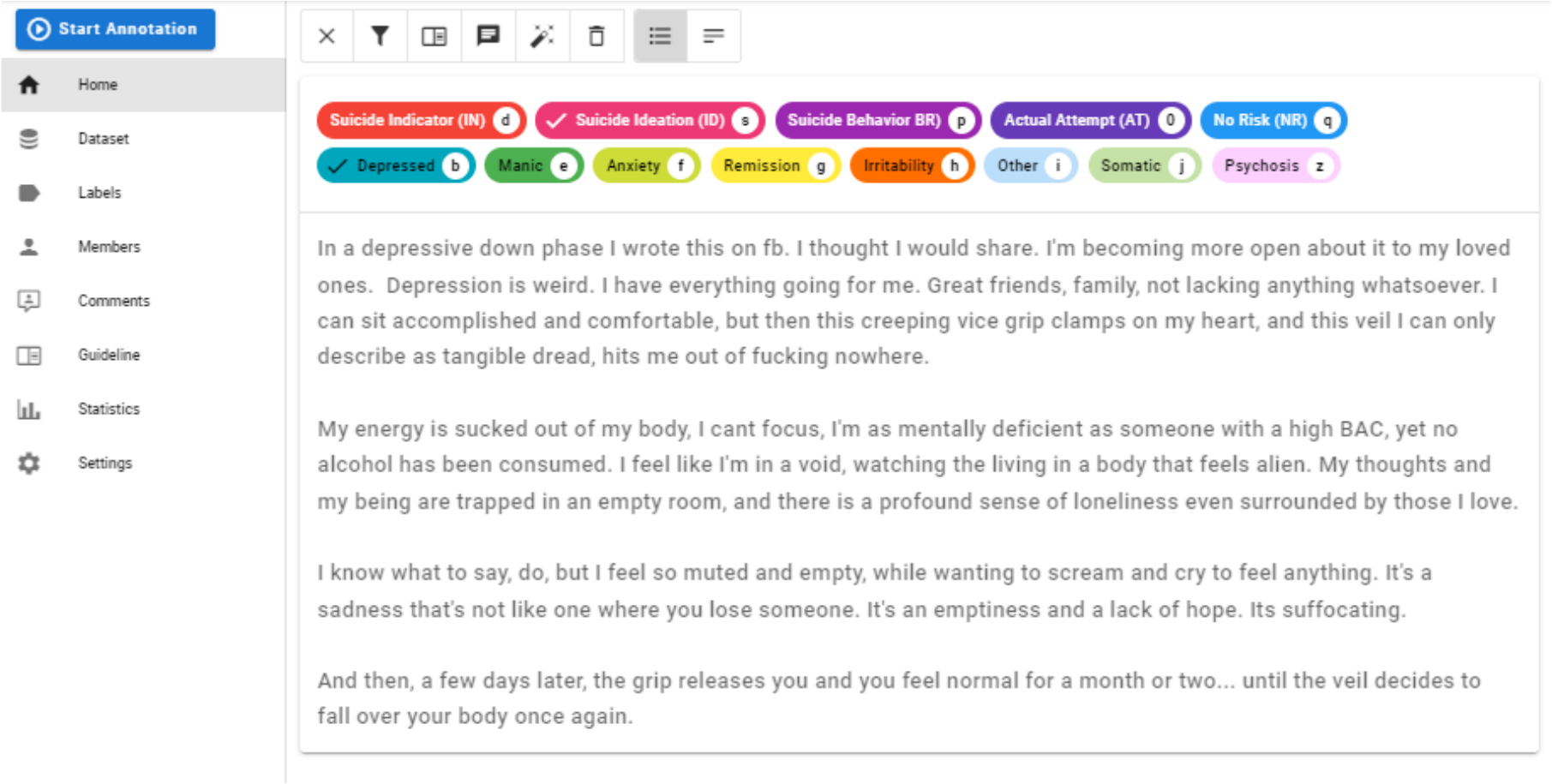}
  \caption{The screenshot of the annotation tool for creating data.}
  \label{fig:doccano}
 \end{figure} 

\subsection{Annotation Process} 
    In this study, we aim to develop a model to predict the risks of future suicidality of bipolar disorder (BD) patients using past social media data. To this end, we create a BD dataset that includes the labels of future suicidality and bipolar symptoms clinically verified by psychiatrists. This section briefly states how we create a BD dataset based on our annotation guideline. As our first step, we collect social media posts published between January 1, 2008, and September 31, 2021, from three bipolar-related subreddits using the open-source \textit{Reddit API} \footnote{\url{https://www.reddit.com/dev/api/}}. Among the collected posts, we used posts written by users who have been diagnosed with BD by professionals~\cite{jagfeld2021understanding} and users who reported BD diagnosis (e.g., ``I was diagnosed with Bipolar type-I last year.''). Based on the criteria, our dataset contains 7,592 posts published by 818 users, i.e., BD patients. For the preprocessing, we anonymize the collected posts and convert the texts. Then we conduct annotation with four trained annotators using the open-source text annotation tool \textit{Doccano} in Figure~\ref{fig:doccano}. 
    
    % 10.13 We aim to build a dataset to develop a deep learning model that predicts future suicidal risk based on a history of bipolar symptoms. As our first step, we collect social media posts related to Bipolar using data from open source Reddit API~\footnote{\url{https://www.reddit.com/dev/api/}}, an online site for anonymous discussion. For the preprocessing, we anonymize the collected posts and remove personally identifiable information. Then we conduct the annotation with the four annotators on an open-source text annotation tool doccano in Figure~\ref{fig:doccano}. 
    
    % 이전 In this project, we aim to build a dataset to develop a deep learning model which predicts future suicide risk based on Bipolar Symptom History. As our first step, we collect social media posts related to Bipolar using data from Reddit, an online site for anonymous discussion. For the preprocessing, personally identifiable information was removed, and content was de-identified. Then we conduct the annotation with the four annotators on an open-source text annotation tool doccano in Figure~\ref{fig:doccano}. 

\subsection{Annotation Guideline} \label{sec:appendix_annot_guide}
    For our annotation, we consider three different label categories that include the diagnosed BD type (e.g., BD-I, BD-II), the BD symptom (e.g., manic, anxiety), and the level of suicidality  (e.g., ideation, attempt). Discussion with the psychiatrist selected the criteria of three different label categories. We briefly describe the details of the annotation guideline in the following subsections.
    % 10.13 In this section, we briefly state the annotation guideline in the following order: \textit{1) Self-report Bipolar Disorder Diagnosis, 2) Bipolar Symptom, and 3) Suicidal Risk.} Discussion with the psychiatrist selected the criteria of three categories.

\subsubsection{Diagnosed Bipolar Disorder Types} \label{sec:appendix_annot_diagnosis}
    To use only posts of users diagnosed with bipolar by medical institutions, we classify users whose self-reports are bipolar related to diagnoses (e.g., ``Hey, I'm diagnosed bipolar II posts being diagnosed with schizophrenia.''). We label users into three BD diagnosis types, including \textit{Bipolar Disorder-I} (BD-I), \textit{Bipolar Disorder-II} (BD-II), and \textit{Not Otherwise Specified Bipolar Disorder} (NOS). Table \ref{tab:keyword} describes the definition of three BD diagnosis types inspired by the Diagnostic and Statistical Manual of Mental Disorders (DSM-5)~\cite{APA2013} and the Statistical Classification of Diseases and Related Health Problems (ICD-10)~\cite{world2016international}, which classify bipolar disorder into several sub-types based on the frequency and intensity of episodes. For example, \textit{BD-I} requires at least one manic episode, while \textit{BD-II} shows at least one hypomanic and one major depressive episode during their lifetime. Moreover, we annotate \textit{NOS} when a patient shows some symptoms of BD but does not necessarily satisfy all the criteria.

\begin{table*}[h!]
\centering
\caption{Definition of diagnosed BD types.}
\label{tab:keyword}
\resizebox{.95\linewidth}{!}{%
\begin{tabular}{p{0.17\linewidth}|p{0.83\linewidth}}
    {\bf BD Type}    &  {\bf Definition} \\ \toprule
    BD-I  & At least one-lifetime manic sepisode and one major depressive episode.\\
    BD-II  & At least one hypomanic and one major depressive episode. \\
    NOS &  Shows some symptoms of BD but does not necessarily satisfy all the criteria. \\ \bottomrule
\end{tabular}}
\end{table*}

    % 10.13 To use only social media posts of users diagnosed with Bipolar by medical institutions, we classified users whose Self-reports are Bipolar related to diagnoses (e.g., `Hey I'm diagnosed bipolar|| (23f) post being diagnosed with schizophrenia (stress-related psychosis).'). Table \ref{tab:keyword} describes the definition of three bipolar diagnosis categories inspired by the Diagnostic and Statistical Manual of Mental Disorders (DSM-5)~\cite{APA2013} and the  Statistical Classification of Diseases and Related Health Problems (ICD-10)~\cite{world2016international}, which classify bipolar disorder into several sub-types based on the frequency and intensity of episodes. For example, \textit{bipolar type I (BD-I)} requires at least one manic episode, while \textit{bipolar type II (BD-II)} is at least one hypomanic and one major depressive episode during their lifetime. Moreover, we annotate \textit{Bipolar Disorder Not Otherwise Specified (NOS)} in two cases: 1) the user's diagnosis was specifically explicit as NOS, and 2) not mentioning a specific bipolar diagnosis.

  %\begin{itemize}[leftmargin=*]
   %     \item \textit{Delete Criteria: } 1) self-diagnoses posts and 2) any posts related to other %individuals, including family members or friends.
 %\end{itemize}
% 효림님 section도 sub, subsubsection까지 하면 알아서 앞에 번호 달아줍니다
% 작은따옴표 안에 글을 넣고 싶을 땐 '' => `' 이렇게 해야합니다! (shift + 물결)
% 영어에서 괄호는 띄어쓰기 해야합니다 예) 한국어 : 다은(오키) /영어: daeun (okay)

\begin{table*}[h]
    \centering
    \caption{The descriptions and examples in Bipolar Disorder symptoms and risks of suicidality}
    \resizebox{.95\linewidth}{!}{%
    \begin{tabular}{p{0.1\linewidth}|p{0.08\linewidth}|p{0.82\linewidth}}
    \textbf{Category}  &  \textbf{Symptom} & \textbf{Description} \& \textbf{Example}  \\ \toprule
    BD Symptom  & Depressed & sadness, feeling of inadequacy, psychomotor slowing, social withdrawal, reduced sex drive, etc \\  
    &  & {\it “I’ve always had unstable moods and antidepressants have only ever made things worse for me.”} \\ 
    \cline{2-3}
      & Manic & elated mood, hyperactive, increased sexuality, risky behavior, impulsive, distractible, etc \\  
    &  & {\it “I have so much adrenaline that I'll start laughing to myself at my own jokes, singing or shouting.”}  \\
    \cline{2-3}
      & Anxiety & anxious mood, somatic anxiety, agitation, sense of nervousness, obsession, etc \\  
    &  & {\it “But what if i’m just having a good week, i get nervous now ... i don’t know what to think or feel.” } \\ \cline{2-3}    
    & Remission & symptom relief, excellent, feeling of comfort, happiness, etc \\  
    &  & {\it "As for how I feel, I feel a massive sense of relief." } \\    \cline{2-3}
      & Irritability & irritable mood, annoyance, impatient, anger, sensitiveness, scream, etc \\  
    &  & {\it “I'm just so annoyed with everyone and everything all I wanna do is scream.”}  \\ 
    \cline{2-3}
      & Somatic & insomnia/hypersomnia, decreased/increased appetite, impaired concentration, amnesia, etc  \\  
    &  & {\it “At first my appetite was normal but now it’s been going away. I feel no hunger during the day.” }  \\  
    \cline{2-3}
      & Psychosis & persecutory idea, delusion, hallucinations, impaired insight, etc \\  
    &  & {\it “I feel paranoid, have been having delusions and I saw people.”  } \\ 
    \midrule
    Suicidality  & Indicator & Risk indicators such as a history of divorce, chronic illness, or suicide of a loved one. \\  
    &  & {\it “I’ve been waking up and crying first thing every day for the past several days.”} \\ 
    \cline{2-3}
      &  Ideation & Any mention of wanting to take one’s own life. \\  
    &  & {\it “I still want to die. i still should die. i feel sorry for anyone who knows me.”}  \\
    \cline{2-3}
      & Behavior & Actions with higher risk such as cutting or planning for a suicide attempt. \\  
    &  & {\it “Imma go cut myself into pieces like I deserve. World is a much better place without me.”}  \\ 
    \cline{2-3}
      & Attempt & Letter asking for help after the suicide attempt, will, deliberate action that can lead to death.  \\  &  & {\it “I failed suicide last night, what do I do now? I thought I took enough pills to kill myself last night.” }  \\  \midrule
    \bottomrule
    \end{tabular}}
    \label{tab:w2v}
\end{table*} 

\subsubsection{Bipolar Disorder Symptoms} \label{sec:appendix_annot_symptom}

    We can filter out posts without BD diagnosis type from Appendix \ref{sec:appendix_annot_diagnosis}. We annotate BD symptoms for each post that fits the requirement to track users diagnosed as bipolar with time series. Based on the BISS(Bipolar Inventory of Symptoms Scale) \citet{bowden2007development} and discussions with the psychiatrist, our annotation criteria came out. We annotate users in case bipolar-related symptoms are exposed. Table ~\ref{tab:w2v} describes the definition and the corresponding examples of BD symptoms used in this study. For more systematic annotation, we consider mood and somatic symptoms. We first annotate the most prominent mood symptoms among \textit{Depressed, Manic, Anxiety, Remission, Irritability} and \textit{Other}. Additionally, we add \textit{Other} to cover moods that do not fall into the other five mood symptoms. Moreover, simultaneously with some posts, we annotate an additional somatic symptom label in option with \textit{Somatic complaint, Psychosis}, and \textit{Both}, which are considered vital factors of suicidality ~\cite{bardram2012monarca,simpson1999risk}. While annotating, we delete advertising posts that do not fit the purpose.
    
    % From Appendix \ref{sec:appendix_annot_diagnosis} we can filter out posts for which BD diagnosis type was not stated. In other words, from this annotation step, each post should consist of 1) cases of bipolar symptoms and 2) suicidal risk levels. Using posts that fit the requirements, we check bipolar symptoms with time series to track users who are diagnosed as bipolar. Based on the BISS(Bipolar Inventory of Symptoms Scale) \citet{bowden2007development} and discussions with the psychiatrist, our annotation criteria came out. We annotate users in case bipolar-related symptoms are exposed. Table ~\ref{tab:w2v} describes the definition and the corresponding examples of bipolar symptom categories used in this paper. For more systematic annotation, bipolar symptoms are divided into \textit{1) Mood} and \textit{2) Somatic}. For \textit{1) Mood}, the most prominent symptom was annotated among \textit{Depressed, Manic, Irritability, Anxiety, and Remission}. Additionally, we add \textit{No} to deal with the rest, which does not correspond to the five moods. Moreover, simultaneously with some posts, \textit{2) Somatic} was annotated in option with \textit{Somatic complaint} and \textit{Psychosis}, which are considered vital suicide risk factors~\cite{bardram2012monarca,simpson1999risk}. While annotating, we delete Advertising posts that do not fit for the purpose.

\subsubsection{Risks of Suicidality} \label{sec:appendix_annot_suicidal_risk}
    To determine the user's different levels of suicidality while tracking BD symptoms, we also simultaneously annotate the risks of suicidality. Based on the post contents, we label the risk of suicidality, which fits the current situation. We utilize the existing criteria from \cite{gaur2019knowledge} that provide five levels of suicidality, including \textit{No Risk}~(NR), \textit{Suicide Indicator}~(IN), \textit{Suicidal Ideation}~(ID), \textit{Suicidal Behavior}~(BR), and \textit{Actual Attempt}~(AT), based on the Columbia Suicide Severity Rating Scale (C-SSRS)~\cite{posner2011columbia}. For our annotation, we merge \textit{No Risk} with \textit{Suicide Indicator} since people with bipolar disorder are already considered more at risk than the general population in suicide~\cite{rihmer2002bipolar,ilgen2010psychiatric}. In the suicide indicator level, posts reveal risk indicators such as a history of divorce, chronic illness, or suicide of a loved one. Suicidal ideation posts mention the willingness to take own life (e.g., ``I still want to die. I still should die.''), and suicidal behavior posts contain actions with higher risks, such as planning a suicide attempt. Posts show deliberate action at the actual attempt level that can lead to death (e.g., “I failed to commit suicide last night, what do I do now?”). Table ~\ref{tab:w2v} details each category's descriptions and examples of Risks of Suicidality.

    % 10.13 To determine the user's suicide risk level while tracking bipolar symptoms, we annotate Suicidal Risk simultaneously. Based on post contents, we label suicide risk, which fits the current situation. The guideline was developed concerning the Columbia Suicide Severity Rating Scale (C-SSRS) written by \citet{gaur2019knowledge} for identifying Suicidal risk on social media platforms. The symptom escalates in the following order: Suicide Indicator (IN), Suicidal Ideation (ID), Suicidal Behavior (BR), and Actual Attempt (AT). Table ~\ref{tab:w2v} presents more details about the descriptions and examples of Suicidal Risk in each category.

\begin{table*}[]
\caption{Differences between target and control groups based on LIWC results and annotated results.}
\label{tab:total T-test}
\begin{tabular}{@{}lcc|lcclcc@{}}
\toprule
\multicolumn{3}{c|}{\textbf{Annotated Result}}           & \multicolumn{6}{c}{\textbf{LIWC}}                                                                                                      \\ \midrule
\multicolumn{1}{c}{\textbf{}}  & \textbf{t} & \textbf{p} & \multicolumn{1}{c}{\textbf{}} & \textbf{t} & \multicolumn{1}{c|}{\textbf{p}} & \multicolumn{1}{c}{\textbf{}} & \textbf{t} & \textbf{p} \\ \midrule
\textbf{Current Suicidality} &            &            & \textbf{Mood}                 &            & \multicolumn{1}{c|}{}           & \textbf{Body}                 &            &            \\
- Indicator                           & -11.91     & 0.000 *    & - posemo                      & 0.96       & \multicolumn{1}{c|}{0.339}      & - percept                     & -1.41      & 0.159      \\
- Ideation                     & 9.50        & 0.000 *    & - negemo                      & 3.51       & \multicolumn{1}{c|}{0.000 *}    & - see                         & -1.71      & 0.087      \\
- Behavior                     & 6.59       & 0.000 *    & - anx                         & -1.47      & \multicolumn{1}{c|}{0.143}      & - hear                        & -1.12      & 0.263      \\
- Attempt                      & 3.39       & 0.001 *    & - anger                       & 2.85       & \multicolumn{1}{c|}{0.004 *}    & - feel                        & 1.24       & 0.216      \\
\textbf{BD Symptom}            &            &            & - sad                         & 5.16       & \multicolumn{1}{c|}{0.000 *}    & - bio                         & 1.70       & 0.089      \\
- Manic                        & -3.88      & 0.000 *    & - death                       & 8.03       & \multicolumn{1}{c|}{0.000 *}    & - body                        & 1.58       & 0.115      \\
- Anxiety                      & -0.36      & 0.722      & \textbf{Social}               &            & \multicolumn{1}{c|}{}           & - health                      & 1.58       & 0.264      \\
- Irritability                 & -6.07      & 0.000 *    & - social                      & -1.84      & \multicolumn{1}{c|}{0.066}      & - achieve                     & -2.59      & 0.010 *     \\
- Remission                    & -0.28      & 0.778      & - family                      & 2.23       & \multicolumn{1}{c|}{0.026 *}    & \textbf{Liguistic}            &            &            \\
- Somatic                      & 1.11        & 0.267        & - friend                      & 0.98       & \multicolumn{1}{c|}{0.325}      & - i                           & 5.73       & 0.000 *    \\
- Psychosis                    & 2.44       & 0.01      & - affiliation                 & -0.90      & \multicolumn{1}{c|}{0.370}       & - we                          & -1.37      & 0.171      \\
                               &            &            & - work                        & -2.38      & \multicolumn{1}{c|}{0.017 *}    & - you                         & -1.88      & 0.060       \\
                               &            &            & - money                       & -2.22      & \multicolumn{1}{c|}{0.027 *}    & - shehe                       & -1.87      & 0.061      \\
                               &            &            & - relig                       & 1.41       & \multicolumn{1}{c|}{0.159}      & - they                        & -0.05      & 0.957      \\
                               &            &            & - risk                        & 2.03       & \multicolumn{1}{c|}{0.042 *}    &                               &            &            \\ \bottomrule
\end{tabular}
\end{table*}

\section{Experiment Settings} \label{sec:appendix_exset}
    We tune hyperparameters based on the highest F1 score obtained from the cross-validation set for the models. We use the grid search to explore the dimension of hidden state $ H \in \{32, 64, 128, 256, 512\} $, number of LSTM layers $ n \in \{1,2,5\}$, dropout $\sigma \in \{0.1, 0.2, 0.3, 0.4, 0.5\} $, initial learning rate $lr \in \{1e-5, 2e-5, 3e-5, 5e-5\}$, and control the parameter for ordinal regression $\alpha \in \{0, 0.2, ..., 3.8\}$. The optimal hyperparameters were found to be: $ H = 512 $,$ n = 2 $, $\sigma = 0.1 $, $lr = 1e-5$, and $\alpha = 1.8$. We implement all the methods using PyTorch 1.6 and optimize with the mini-batch AdamW~\cite{loshchilov2017decoupled} with a batch size of 64. We use the Exponential Learning rate Scheduler with gamma $0.001$. We train the model on a GeForce RTX 2080 Ti GPU for 200 epochs and apply early stopping with patience of 20 epochs.

\end{document}